\documentclass[A4,preprint,12pt,number]{elsarticle}

\usepackage{amssymb}
\usepackage{amsmath}

\usepackage{subcaption}
\usepackage{graphicx} 
\usepackage{booktabs}
\usepackage{algorithm}
\usepackage{algpseudocode}
\usepackage{xcolor}
\usepackage{subcaption}
\usepackage[colorlinks=true, linkcolor=blue, citecolor=blue, urlcolor=blue]{hyperref}
\usepackage{float}  
\usepackage{algorithm}
\usepackage{multirow}
\usepackage{algpseudocode}

\journal{}

\begin{document}

\begin{frontmatter}



\title{FlatFormer: A Flat Transformer Knowledge Tracing Model Based on Cognitive Bias Injection}

\cortext[cor1]{Corresponding author.}

\author[1]{Xiao-li Xia}

\author[1]{Hou-biao Li\corref{cor1}}
\ead{lhb0189@uestc.edu.cn}

\affiliation[1]{
    organization={School of Mathematical Sciences, University of Electronic Science and Technology of China}, 
    addressline={No. 2006, Xiyuan Avenue, West Hi-Tech Zone}, 
    city={Chengdu},
    postcode={611731}, 
    state={Sichuan},
    country={China}
}

\begin{abstract}
Knowledge Tracing (KT) models face a critical ``Performance-Complexity Trap'': capturing complex cognitive dynamics like learning sessions and memory decay typically requires deep hierarchical architectures, which incur prohibitive computational costs for real-time deployment. To resolve this, we propose \textbf{FlatFormer}, a streamlined architecture based on the novel design paradigm of ``Information Injection over Structural Stacking.'' Unlike parameter-heavy hierarchical models, \textbf{FlatFormer} leverages a standard flat Transformer augmented with two lightweight injection mechanisms: (i) a hybrid input encoding strategy combining learnable session identifiers with fixed sinusoidal step embeddings; and (ii) a pre-computed power-law bias integrated directly into attention logits to explicitly model the forgetting curve. Extensive experiments on four large-scale datasets (e.g., EdNet, Junyi) show that \textbf{FlatFormer} achieves state-of-the-art performance. For example, on the EdNet dataset, compared to the strongest hierarchical baseline (HiTSKT), its absolute AUC increased by \textbf{8.3\%}, while using less than \textbf{15\%} of parameters, and inference speed was about \textbf{three times} faster. These results validate that high cognitive fidelity does not necessitate architectural complexity.
\end{abstract}




\end{frontmatter}



\section{Introduction}
\label{sec:intro}

The exponential scalability of online education platforms, such as Coursera and Khan Academy, has placed Intelligent Tutoring Systems (ITS) at the forefront of modern educational technology. As critical expert systems, the primary function of an ITS is to provide personalized learning paths and instant feedback to millions of concurrent learners~\cite{hooshyar2022review, feng2009addressing}. Central to this capability is Knowledge Tracing (KT)~\cite{liu2021survey, dkt2015}, a technique that dynamically tracks a student's evolving mastery state. However, as user bases expand to hundreds of millions of interactions (e.g., EdNet~\cite{ednet2020}), the practical deployment of KT models faces a severe engineering bottleneck: \textbf{real-time responsiveness}. An effective expert system must not only predict student performance accurately, but must do so with minimal latency to ensure a seamless interactive experience, potentially even on resource-constrained edge devices.

Despite this requirement for efficiency, the recent trajectory of Knowledge tracking research has increasingly drifted towards a ``Performance-Complexity Trap.'' While early models like DKT~\cite{dkt2015} treated learning history as simplified flat sequences---ignoring complex cognitive dynamics such as learning sessions and memory decay~\cite{tskt2021, ebbinghaus1913memory}---recent State-of-the-Art (SOTA) solutions have swung to the opposite extreme. Adopting a philosophy of ``Structural Stacking'' models such as HiTSKT~\cite{hitskt2021}, GKT~\cite{gkt2019}, and TSKT~\cite{tskt2021} employ deep hierarchical encoders and complex graph neural networks to explicitly model session structures and knowledge topology. Although these ``heavyweight'' architectures improve predictive accuracy, they incur a prohibitive increase in computational cost and inference latency. This results in a widening ``performance-efficiency gap'' rendering such parameter-heavy models impractical for large-scale real-time industrial deployment.

\begin{figure*}[t]
    \centering
    \begin{subfigure}[b]{0.495\textwidth}
        \centering
        \includegraphics[width=\linewidth]{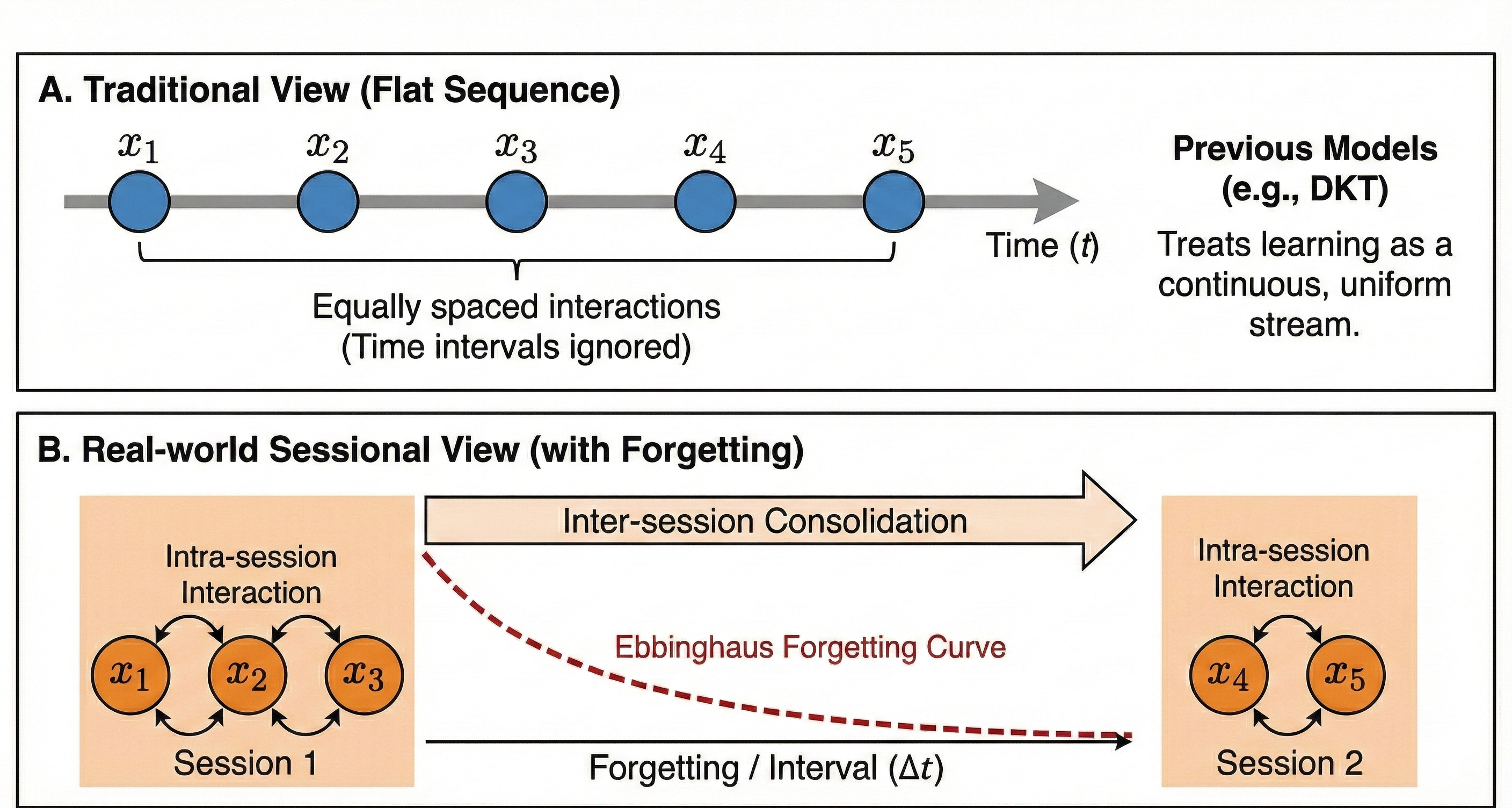} 
        \caption{} 
        \label{fig:motivation_a}
    \end{subfigure}
    \hfill 
    \begin{subfigure}[b]{0.495\textwidth}
        \centering
        \includegraphics[width=\linewidth]{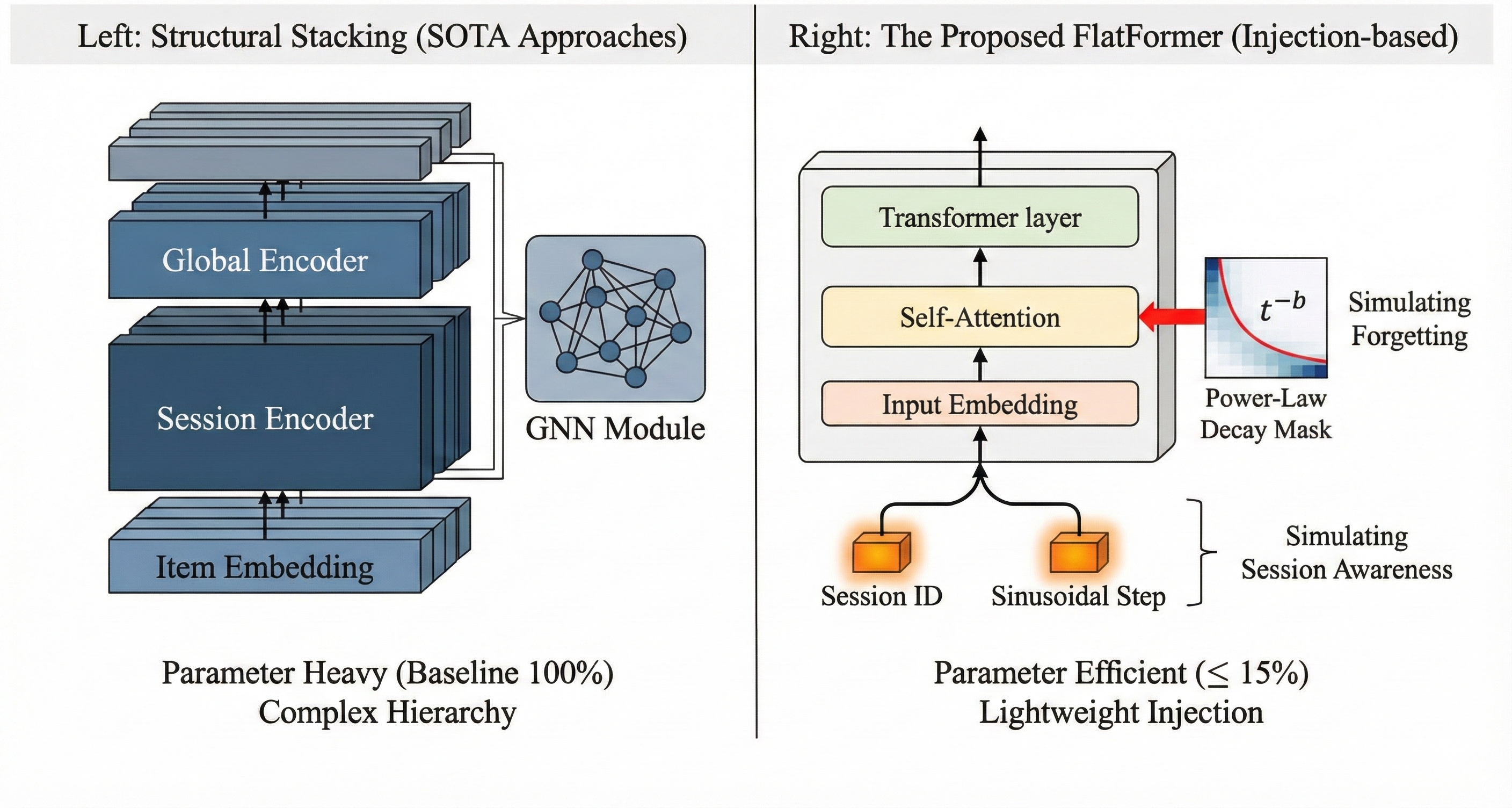}
        \caption{} 
        \label{fig:motivation_b}
    \end{subfigure}
    
    \caption{Conceptual comparison between simplified sequence assumptions and hierarchical cognitive processes. \textbf{(a)} Cognitive dynamics illustrating intra-session interactions ($\tau_t$) and inter-session consolidation ($s_t$). \textbf{(b)} Comparison of hierarchical architectures (left) with the proposed FlatFormer framework (right).}
    \label{fig:intro_concept}
\end{figure*}

In this work, we propose \textbf{FlatFormer}, a novel framework designed to resolve this engineering dilemma by adhering to the general design paradigm of ``Information Injection over Structural Stacking''. Specifically, we instantiate this paradigm via a strategy we term \textbf{Cognitive Injection}: we posit that advanced cognitive features---specifically session awareness (as conceptualized in Figure~\ref{fig:intro_concept}(a), where interactions correspond to $\tau_t$ and consolidation to $s_t$) and power-law forgetting---can be efficiently modeled within a standard, flat Transformer architecture by strategically injecting inductive biases rather than stacking complex layers.

As illustrated in Figure~\ref{fig:intro_concept}(b), we depart from hierarchical designs and introduce two lightweight injection mechanisms:
\begin{itemize}
    \item \textbf{Input Layer Injection (hereafter referred to as Injection-i):} To replace computationally expensive hierarchical encoders, we inject a hybrid encoding scheme. This combines learnable ``Session IDs'' to delineate distinct learning sessions with fixed sinusoidal ``Within-Session Step'' embeddings, effectively capturing cross-session dependencies with minimal overhead.
    \item \textbf{Attention Layer Injection (hereafter referred to as Injection-ii):} Instead of utilizing complex gated units or temporal modules, we inject a pre-computed, non-parametric ``Power-Law Bias'' directly into the self-attention logits. This mechanism explicitly attenuates attention towards distant historical interactions, simulating memory decay without introducing additional learnable parameters.
\end{itemize}

By prioritizing information injection, FlatFormer achieves robust cognitive modeling with a minimal architectural footprint. Extensive experiments on four real-world datasets demonstrate that FlatFormer not only matches the predictive accuracy of hierarchical baselines but does so while utilizing less than $15\%$ of their parameters and achieving significantly faster inference speeds.

The main contributions of this study are summarized as follows:
\begin{enumerate}
    \item We propose FlatFormer, a streamlined architecture that validates the ``Information Injection'' paradigm, demonstrating that standard Transformers can efficiently capture hierarchical cognitive structures without the computational overhead of structural stacking.
    \item We introduce two lightweight mechanisms, hybrid session encoding and attention logit decay, that effectively substitute complex hierarchical encoders, proving that cognitive fidelity does not require parameter redundancy.
    \item Empirical evaluations across four benchmark datasets confirm that FlatFormer achieves SOTA-comparable performance while reducing parameter count by over $85\%$ and increasing inference speed by approximately $3$ times, offering a highly efficient solution for practical, real-time ITS deployment.
\end{enumerate}
\section{Related Work}
\label{sec:related}

\subsection{Deep Sequence Modeling for Knowledge Tracing}
The paradigm shift from probabilistic methods (e.g., Bayesian Knowledge Tracing) to deep learning has fundamentally advanced the field of Knowledge Tracing (KT). \textbf{Deep Knowledge Tracing (DKT)}~\cite{dkt2015} initiated this transition by utilizing Long Short-Term Memory (LSTM) networks to capture the continuous evolution of student knowledge states. To mitigate the inherent limitations of RNNs—specifically, limited receptive fields and difficulty in parallelization—attention-based mechanisms were introduced. Models such as \textbf{SAKT}~\cite{sakt2019} and \textbf{SAINT}~\cite{saint2020} leverage self-attention to identify long-range dependencies between interactions. Subsequent advancements like \textbf{AKT}~\cite{akt2020} incorporate context-aware attention with monotonic assumptions to implicitly model the decay of knowledge over time.

Despite their empirical success, these general-purpose sequence models typically conceptualize student history as a singular, continuous ``flat sequence''. This simplification suffers from \textit{temporal oversimplification}: it fails to distinguish between immediate working memory (intra-session) and long-term consolidation (inter-session), and it lacks explicit mechanisms to quantify the forgetting curve, often relying on the model's implicit capacity to learn these dynamics from massive data.

\subsection{Structural Stacking for Cognitive Dynamics}
To overcome the limitations of flat sequence modeling, recent literature has increasingly adopted a philosophy of ``Structural Stacking,'' where specialized architectural modules are added to capture specific cognitive nuances. We categorize these approaches based on the cognitive dynamic they target:

\begin{itemize}
    \item \textbf{Modeling Session Awareness:} Recognizing that learning occurs in discrete intervals, \textbf{HiTSKT}~\cite{hitskt2021} introduces a hierarchical Transformer framework. It utilizes two distinct encoders—one for intra-session interactions and another for inter-session evolution—to explicitly model knowledge consolidation. While effective, this dual-encoder design significantly increases parameter complexity and inference latency.

    \item \textbf{Modeling Forgetting Mechanisms:} To capture memory decay, \textbf{TSKT} ~\cite{tskt2021} integrates a continuous-time decay kernel directly into the attention mechanism, while \textbf{HawkesKT}~\cite{hawkeskt2022} models the temporal excitement of learning events using Hawkes processes. These methods often require complex numerical solvers or specialized temporal embedding modules, linearly increasing the computational burden with the temporal span of the user history.

    \item \textbf{Modeling Knowledge Relations:} Approaches such as \textbf{GKT}~\cite{gkt2019} and \textbf{SKT}~\cite{skt2021} incorporate Graph Neural Networks (GNNs) to aggregate prerequisite dependencies from concept graphs. Although these graph-based enhancements improve interpretability, they necessitate the maintenance and processing of large adjacency matrices, further complicating the model architecture.
\end{itemize}
\subsection{The Efficiency-Accuracy Dilemma}
The pursuit of higher cognitive fidelity has inadvertently created a ``performance complexity trap''. As detailed above, State-of-the-Art (SOTA) improvements are predominantly achieved through architectural expansion. For instance, the hierarchical design of HiTSKT requires maintaining separate parameters for different temporal granularities, resulting in a model size $>7\times$ that of standard attention baselines (e.g., SAKT). Even recent streamlined attempts, such as \textbf{SimpleKT}~\cite{simplekt2022}, focus on embedding simplification rather than addressing the structural overhead required for complex temporal modeling.

Consequently, there is a scarcity of methods capable of capturing advanced cognitive dynamics (specifically sessions and forgetting) without sacrificing the parameter economy of a standard backbone. \textbf{FlatFormer} addresses this gap by shifting the paradigm from ``Structural Stacking'' to ``Information Injection'', demonstrating that explicit cognitive biases can be embedded into a flat architecture to achieve SOTA performance with a fraction of the computational cost.
\section{Problem Formulation}
\label{sec:problem_formulation}

In this section, we formally define the Knowledge Tracing task. We first present the standard formulation used in sequence-based models. Subsequently, we introduce our \textit{Augmented Cognitive Context} formulation, which derives hierarchical cognitive features (session and time-decay) from raw logs to support the proposed injection-based paradigm without altering the fundamental sequence structure.

\subsection{Standard Knowledge Tracing Paradigm}
\label{subsec:problem_standard}

Let $\mathcal{Q} = \{q_1, q_2, \ldots, q_{|Q|}\}$ denote the set of distinct exercises. The learning history of a student is represented as a chronologically ordered sequence $I = \{x_1, x_2, \ldots, x_L\}$, where $L$ is the sequence length. In the standard deep KT paradigm (e.g., DKT, SAKT \cite{dkt2015, sakt2019}), each interaction $x_t$ is typically defined as a tuple $x_t = (e_t, r_t)$, containing:
\begin{itemize}
    \item $e_t \in \mathcal{Q}$: The exercise attempted by the student at step $t$.
    \item $r_t \in \{0, 1\}$: The correctness of the response (1 for correct, 0 for incorrect).
\end{itemize}

The objective of the KT model is to estimate the probability of the student correctly answering the next exercise $e_{L+1}$, given the historical sequence $I$:
\begin{equation}
\label{eq:standard_kt}
P(r_{L+1}=1 \mid e_{L+1}, I).
\end{equation}
This formulation treats $I$ as a ``flat sequence'', implicitly assuming uniform temporal intervals and continuous engagement, thereby neglecting the discrete nature of learning sessions and memory decay.

\subsection{Augmented Cognitive Context Formulation}
\label{subsec:problem_injection}

To resolve the ``efficiency-accuracy dilemma'' identified in Section~\ref{sec:related}, we depart from the ``Structural Stacking'' approach—which restructures $I$ into hierarchical nested lists (e.g., $S = \{ses_1, \ldots, ses_N\}$)—and instead adopt an \textbf{Information Injection} strategy. We posit that cognitive dynamics should be encoded as \textit{features} injected into a flat architecture rather than requiring complex architectural modifications.

We extend the raw interaction tuple to include a timestamp: $x_t = (e_t, r_t, ts_t)$. Based on $ts_t$, we mathematically derive two sets of cognitive context features:

\subsubsection{Derivation of Sessional Features (Session Awareness)}
To address ``sessional blindness'', we explicitly model the boundaries between learning blocks. A new session is defined when the time gap between consecutive interactions exceeds a threshold $\Delta_{gap}$ (typically 30 minutes). We define a mapping function $\Phi_{session}: I \rightarrow (\mathbf{s}, \boldsymbol{\tau})$ that generates two discrete identifiers for each step $t$:
\begin{itemize}
    \item \textbf{Session ID ($s_t \in \mathbb{N}$):} A global counter incremented whenever $ts_t - ts_{t-1} > \Delta_{gap}$. This identifies the specific learning session.
    \item \textbf{Within-Session Step ($\tau_t \in \mathbb{N}$):} A local counter representing the index of the interaction within the current session $s_t$, reset to 1 at the start of each new session.
\end{itemize}
These features $\{s_t, \tau_t\}$ serve as inputs for the \textbf{Injection-i} mechanism.These features $\{s_t, \tau_t\}$ serve as inputs for the \textbf{Injection-i} mechanism.

\subsubsection{Derivation of Temporal Features (Forgetting Awareness)}
To address ``forgetting blindness'', we explicitly model the time intervals that drive memory decay. We define the temporal lag matrix $\mathbf{\Delta T} \in \mathbb{R}^{L \times L}$, where each entry $\Delta t_{t,j}$ represents the elapsed time between the current step $t$ and a historical step $j$ ($j < t$):
\begin{equation}
\label{eq:time_lag}
\Delta t_{t,j} = ts_t - ts_j.
\end{equation}
This matrix $\mathbf{\Delta T}$ serves as the direct input for the \textbf{Injection-ii} mechanism, allowing the model to apply a non-learnable power-law decay mask to the attention scores.

\subsection{Formal Problem Statement}
\label{subsec:problem_statement}

Given the augmented sequence $I_{aug} = \{(e_t, r_t, s_t, \tau_t, ts_t)\}_{t=1}^L$, the goal of \textbf{FlatFormer} is to predict $P(r_{L+1}=1 \mid e_{L+1}, I_{aug})$. 

Crucially, unlike hierarchical models that process $s_t$ and $\tau_t$ using separate encoders, FlatFormer processes $I_{aug}$ as a single flat sequence. The cognitive complexity is handled via the injected features $\{s_t, \tau_t\}$ and $\Delta t_{t,j}$, maintaining the inference efficiency of standard Transformers ($O(L^2)$ or linear variants) while capturing high-fidelity cognitive dynamics. Key notations are summarized in Table~\ref{tab:notations}.

\begin{table}[h!]
\centering
\caption{Summary of Key Notations and Definitions.}
\label{tab:notations}
\begin{tabular}{l p{0.65\linewidth}}
\toprule
\textbf{Notation} & \textbf{Description} \\
\midrule
$I, L$ & The interaction sequence and its length. \\
$\mathcal{Q}, e_t$ & The set of exercises and the specific exercise at step $t$. \\
$r_t \in \{0, 1\}$ & The student's response correctness at step $t$. \\
$ts_t$ & The timestamp of the interaction at step $t$. \\

\addlinespace[1.5ex] 
\multicolumn{2}{l}{\textit{Derived Features}} \\ 
$s_t \in \mathbb{N}$ & \textbf{Session ID:} Identifies the learning session index. \\
$\tau_t \in \mathbb{N}$ & \textbf{Session Step:} The relative index within a session. \\
$\Delta t_{t,j}$ & \textbf{Time Lag:} Elapsed time between step $t$ and $j$, used for forgetting modeling. \\

\addlinespace[1.5ex] 
\multicolumn{2}{l}{\textit{Model Output}} \\
$P(r_{L+1} \mid \cdot)$ & Predicted probability of correctness for the next exercise. \\
\bottomrule
\end{tabular}
\end{table}

\section{The FlatFormer Architecture}
\label{sec:model}

\begin{figure*}[htbp] 
    \centering
    \includegraphics[width=\textwidth]{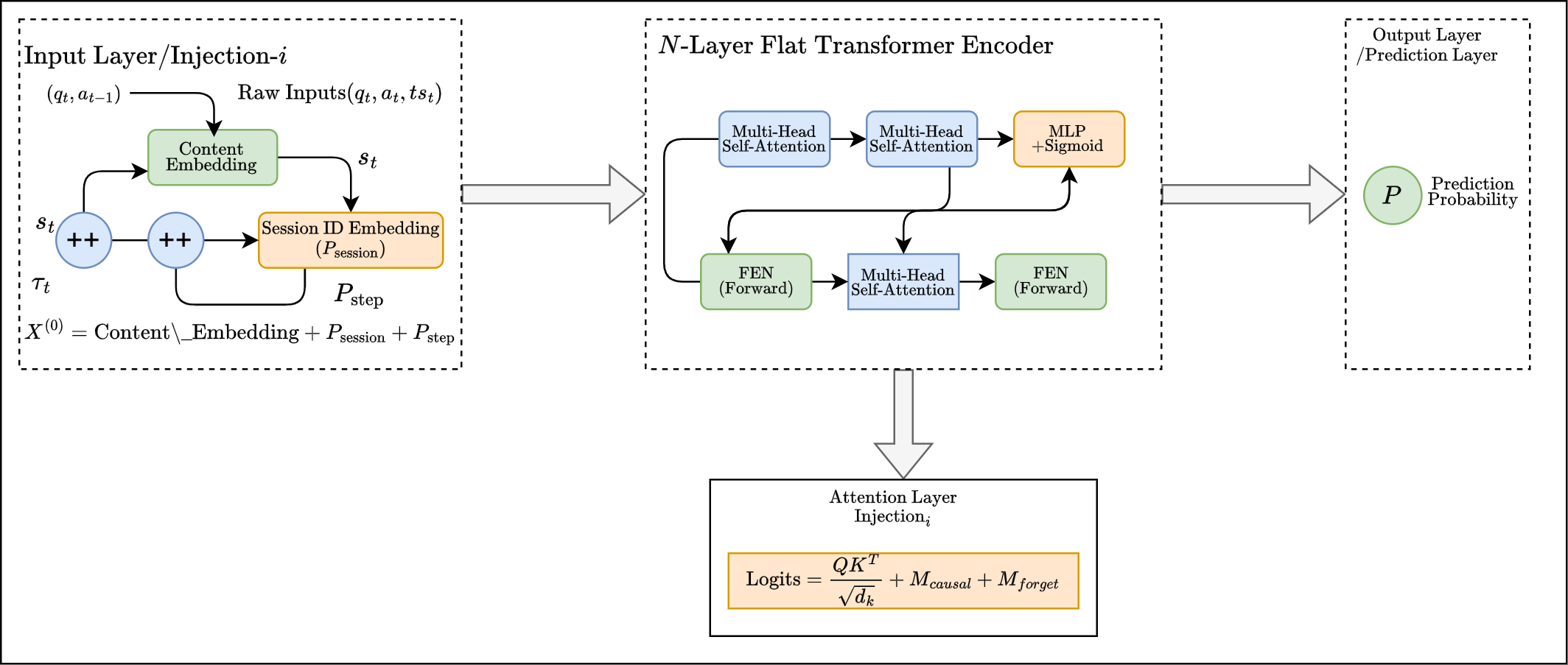} 
    \caption{\textbf{High-level architecture of FlatFormer.} Unlike hierarchical approaches, FlatFormer utilizes a standard flat encoder injected with (1) \textit{Session Features} at the input level and (2) a \textit{Forgetting Bias} at the attention level to efficiently model cognitive processes.}
    \label{fig:model_architecture} 
\end{figure*}
To address the performance-complexity trade-off established in Section~\ref{sec:related} and to instantiate the Cognitive Injection formulation defined in Section~\ref{sec:problem_formulation}, we propose \textbf{FlatFormer}. The core design philosophy of FlatFormer is centered on \textit{Information Injection over Structural Stacking}.

As illustrated in Figure~\ref{fig:model_architecture}, FlatFormer is structured around a standard, flat $N$-layer Transformer Encoder~\cite{vaswani2017transformer}. This contrasts sharply with the deep Hierarchical Encoders adopted by contemporary models such as HiTSKT~\cite{hitskt2021}. The core premise of our design is that modeling cognitive complexity does not necessitate an equivalent degree of architectural complexity.

We concentrate all innovations at two specific cognitive injection points: (i) injecting session features at the input embedding layer (\textbf{Injection-i}) to mitigate sessional blindness; and (ii) injecting a forgetting bias onto the attention logits (\textbf{Injection-ii}, adapting the mechanism from~\cite{press2022alibi}) to address forgetting blindness.

\begin{algorithm}[t!]
\caption{FlatFormer Forward Pass (Part I)}
\label{alg:flatformer_part1}
\begin{algorithmic}[1]
    \State \textbf{Input:} Interaction sequence $I = \{x_1, \ldots, x_L\}$, $x_t = (q_t, a_t, ts_t)$
    \State \textbf{Input:} Hyperparameters $\beta$ (forgetting rate), $\Delta_{gap}$ (session gap)
    \State \textbf{Parameters:} $d_k$, $E_Q, E_A, E_S$, $\{W_l^Q, W_l^K, W_l^V, W_l^O, \text{FFN}_l\}_{l=1}^N$, $\text{MLP}$
    \State \textbf{Output:} Predicted probabilities $P = \{p_1, \ldots, p_L\}$

    \Statex \Comment{\textit{--- 1. (Pre-computation) Feature Derivation (Sec \ref{subsec:problem_injection}) ---}}
    \State $S \gets [ \ ], \tau \gets [ \ ]$ \Comment{Init session \& step lists}
    \State $s_{id} \gets 0, \tau_{step} \gets 0, ts_{\text{last}} \gets 0$
    \For{$t \gets 1$ to $L$}
        \If{$ts_t - ts_{\text{last}} > \Delta_{gap}$}
            \State $s_{id} \gets s_{id} + 1$; $\tau_{step} \gets 0$
        \EndIf
        \State $S.\text{append}(s_{id})$
        \State $\tau.\text{append}(\tau_{step})$
        \State $ts_{\text{last}} \gets ts_t$; $\tau_{step} \gets \tau_{step} + 1$
    \EndFor

    \Statex \Comment{\textit{--- 2. (Pre-computation) Injection-ii: Forgetting Bias (Sec \ref{subsec:model_injection_ii}) ---}}
    \State $M_{\text{forget}} \gets \text{zeros}(L, L)$
    \State $\Delta t_{\max} \gets \max(ts_L - ts_1, 1.0)$
    \For{$t \gets 1$ to $L$}
        \For{$j \gets 1$ to $t$}
            \State $\Delta t_{t,j} \gets ts_t - ts_j$
            \State $\Delta t'_{t,j} \gets \Delta t_{t,j} / \Delta t_{\max}$
            \State $M_{\text{forget}}[t, j] \gets - \beta \cdot \log(\Delta t'_{t,j} + 1)$
        \EndFor
    \EndFor
    \State $M_{\text{causal}} \gets \text{CausalMask}(L)$

    \Statex \Comment{\textit{--- 3. Injection-i: Input Embedding (Sec \ref{subsec:model_injection_i}) ---}}
    \State $E_{\text{content}} \gets E_Q([q_1, \ldots, q_L]) + E_A([a_0, \ldots, a_{L-1}])$
    \State $P_{\text{session}} \gets E_S(S)$
    \State $P_{\text{step}} \gets PE(\tau)$
    \State $X^{(0)} \gets E_{\text{content}} + P_{\text{session}} + P_{\text{step}}$
    
    \Statex \Comment{\textit{(Continued in Algorithm \ref{alg:flatformer_part2})}}
    
    \algstore{flatformer_break}
\end{algorithmic}
\end{algorithm}



\addtocounter{algorithm}{-1}

\begin{algorithm}[t!]
\caption{FlatFormer Forward Pass (Part II)}
\label{alg:flatformer_part2}
\begin{algorithmic}[1]
    \algrestore{flatformer_break}

    \Statex \Comment{\textit{--- 4. FlatFormer Encoder Backbone (Sec \ref{subsec:model_encoder}) ---}}
    \For{$l \gets 1$ to $N$}
        \State $X' \gets \text{LayerNorm}(X^{(l-1)})$
        \State $Q, K, V \gets X'W_l^Q, X'W_l^K, X'W_l^V$
        \State $A \gets \frac{QK^T}{\sqrt{d_k}}$
        \State $A' \gets A + M_{\text{causal}} + M_{\text{forget}}$ \Comment{\textbf{<-- Injection-ii here}}
        \State $H \gets \text{Softmax}(A')V$
        \State $H \gets X^{(l-1)} + \text{MultiHeadConcat}(H)W_l^O$
        \State $X^{(l)} \gets \text{LayerNorm}(H + \text{FFN}_l(H))$
    \EndFor

    \Statex \Comment{\textit{--- 5. Prediction Layer (Sec \ref{subsec:model_prediction}) ---}}
    \State $X^{(N)} \gets \text{LayerNorm}(X^{(N)})$
    \State $P \gets \sigma(\text{MLP}(X^{(N)}))$
    \State \textbf{return} $P$
\end{algorithmic}
\end{algorithm}

\begin{figure}[!t] 
    \centering
    \includegraphics[width=\linewidth]{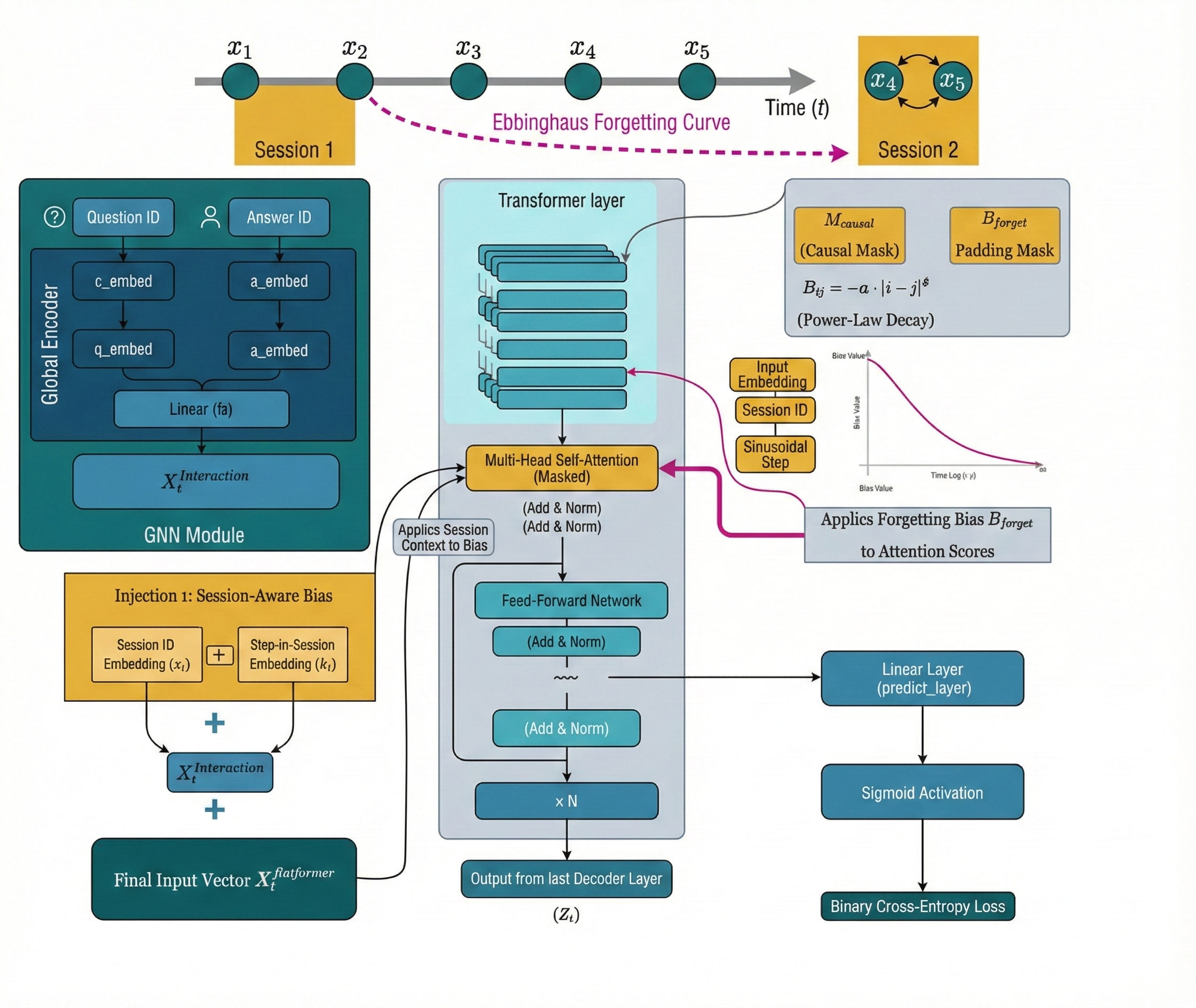} 
    \caption{\textbf{The FlatFormer Model Architecture.} This diagram illustrates the overall workflow, highlighting the two key injection points: (i) \textbf{Session-Awareness} at the Input Layer (Injection-i), and (ii) \textbf{Forgetting Bias} within the Attention Layer (Injection-ii). The model processes raw interactions to predict future student performance.}
    \label{fig:detailed_framework} 
\end{figure}

\textbf{Algorithm~\ref{alg:flatformer_part1}}, visually detailed in Figure~\ref{fig:detailed_framework}, summarizes the full forward pass. The following subsections provide a comprehensive mathematical and conceptual elaboration of each component.

\subsection{Session-Aware Input Embedding Module (Injection-i)}
\label{subsec:model_injection_i}

The function of this module is to transform the discrete interaction tuple $x_t=(q_t, a_t, ts_t)$ and the derived context features $\{s_t, \tau_t\}$ (defined in Sec. \ref{subsec:problem_injection}) into a $d$-dimensional, context-rich vector $X^{(0)}_t$ that serves as the initial input to the encoder stack.

\begin{figure}[t!]
    \centering
    \includegraphics[width=0.91\linewidth]{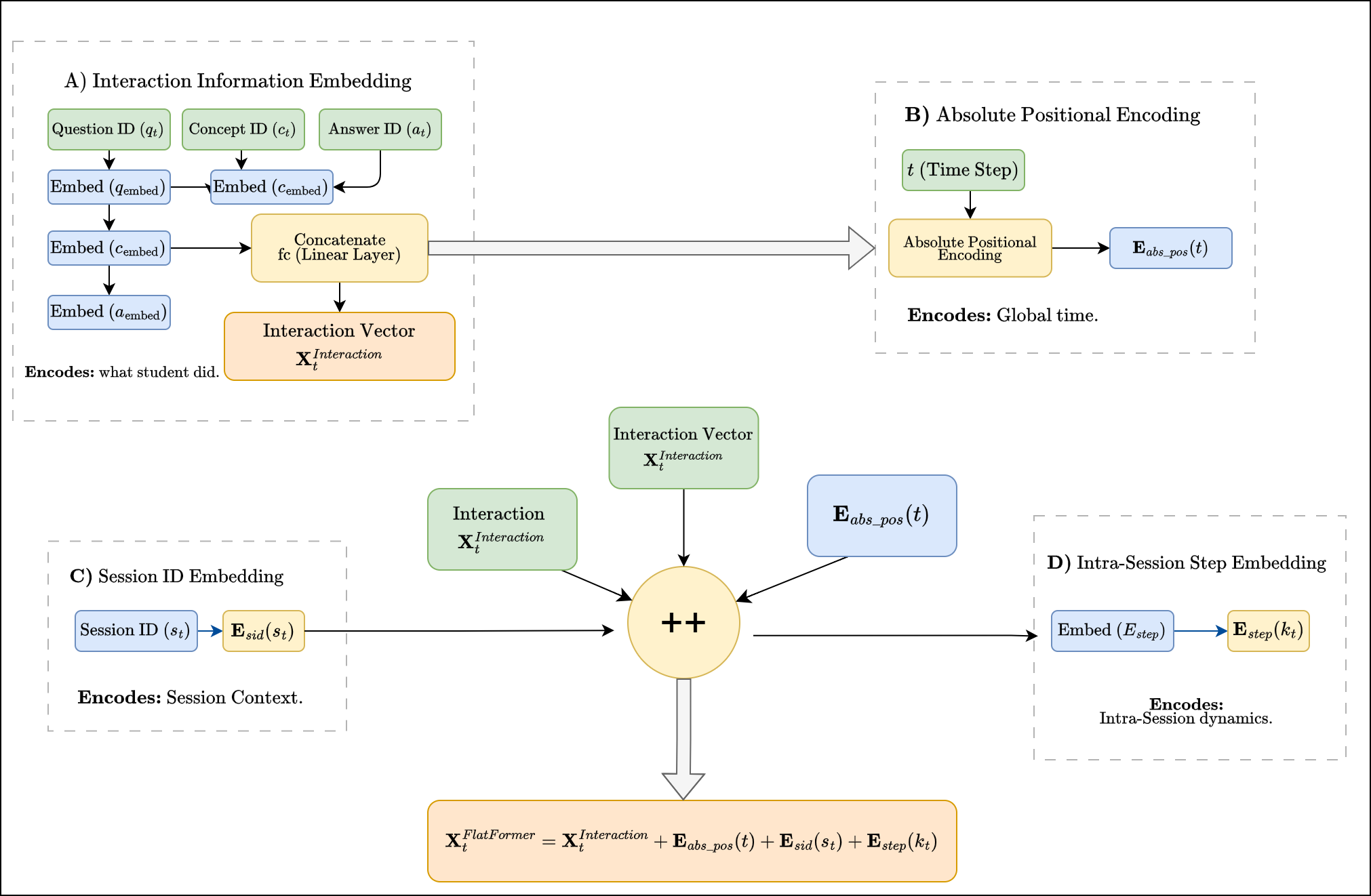} 
    \caption{\textbf{The Architecture of Injection-i.} The model constructs a session-aware input representation by aggregating content embeddings ($E_{content}$), a learnable session ID embedding ($P_{session}$), and a fixed frequency-based step encoding ($P_{step}$). This design directly addresses sessional blindness while enabling infinite extrapolation.}
    \label{fig:injection_i}
\end{figure}

In contrast to standard Transformer inputs (e.g., SAKT~\cite{sakt2019}) that utilize session-agnostic absolute positional encodings (which result in sessional blindness), our approach replaces the conventional positional embedding. Instead, we introduce a composite injection mechanism constructed from three distinct components, as illustrated in Figure \ref{fig:injection_i}.

\subsubsection{Content Embedding}
To capture the semantic content of the interaction, we follow the established paradigm in Knowledge Tracing by incorporating both the current exercise $q_t$ and the student's response to the previous exercise $a_{t-1}$. This design reflects the dependency of the current knowledge state on prior performance. Specifically, the content embedding $E_{\text{content}}(t)$ is computed as the element-wise summation of the exercise and answer embeddings, as shown in Equation (\ref{eq:content_embedding}):
\begin{equation}
E_{\text{content}}(t) = E_Q(q_t) + E_A(a_{t-1}).
\label{eq:content_embedding} 
\end{equation}
where $E_Q \in \mathbb{R}^{|Q| \times d}$ and $E_A \in \mathbb{R}^{2 \times d}$ denote the learnable embedding matrices for exercises and answers, respectively. For the initial time step $t=1$, $a_0$ is initialized as a designated $\langle \text{START} \rangle$ token.

\subsubsection{Session-Awareness Encoding}
To address the "sessional blindness" inherent in standard positional encodings, we introduce a session-awareness module that explicitly injects context via two distinct encoders for the derived features $s_t$ and $\tau_t$.

\textbf{Session ID ($s_t$) Encoding.} The session identifier $s_t$ is modeled as a categorical variable because distinct session IDs lack inherent numerical continuity. Consequently, we employ a learnable embedding matrix $E_S \in \mathbb{R}^{L_S \times d}$ to map each discrete session ID to a dense vector representation:
\begin{equation}
    P_{\text{session}}(t) = E_S(s_t)
\end{equation}
where $L_S$ denotes the maximum number of sessions. This learned representation captures latent inter-session patterns, such as the warm-up effect observed at the beginning of a new study session \cite{lpkt2021}.

\textbf{Within-Session Step ($\tau_t$) Encoding.} In contrast to session IDs, the within-session step $\tau_t$ carries strong ordinal information. While learnable embeddings are a common choice for sequential data, they suffer from two limitations in this context: (1) increased parameter complexity ($L_{\tau} \times d$), and (2) poor generalization to sequence lengths exceeding those seen during training (Out-of-Vocabulary issues). 

To enable infinite extrapolation and minimize parameter overhead, we adopt the non-learnable, frequency-based sinusoidal Positional Encoding (PE) \cite{vaswani2017transformer}:
\begin{equation}
    P_{\text{step}}(t) = PE(\tau_t).
\end{equation}
Formally, for a dimension index $k \in [0, d/2 - 1]$, the encoding is defined as:
\begin{equation}
\begin{cases}
    PE(\tau_t, 2k) = \sin(\tau_t / 10000^{2k/d}), \\
    PE(\tau_t, 2k+1) = \cos(\tau_t / 10000^{2k/d}).
\end{cases}
\end{equation}
This zero-parameter approach effectively encodes relative positions within a session, ensuring the model remains robust to varying session lengths.
\subsubsection{Final Input Representation}
The resultant input representation $X^{(0)}_t$, which serves as the initial input to the Transformer encoder, is obtained via the element-wise summation of the three aforementioned components:
\begin{equation}
    X^{(0)}_t = E_{\text{content}}(t) + P_{\text{session}}(t) + P_{\text{step}}(t).
\end{equation}
This streamlined injection mechanism provides a significant efficiency advantage over prior hierarchical approaches like HiTSKT \cite{hitskt2021}. By substituting the computationally expensive hierarchical encoder with a single lightweight learnable embedding ($E_S$) and a parameter-free frequency encoding ($PE$), our model effectively captures interaction semantics, session context, and intra-session sequential dependencies with minimal architectural complexity.

\subsection{Forgetting-Aware Self-Attention Mechanism}
\label{subsec:model_encoder}

\begin{figure*}[t!] 
    \centering
    \includegraphics[width=0.93\textwidth]{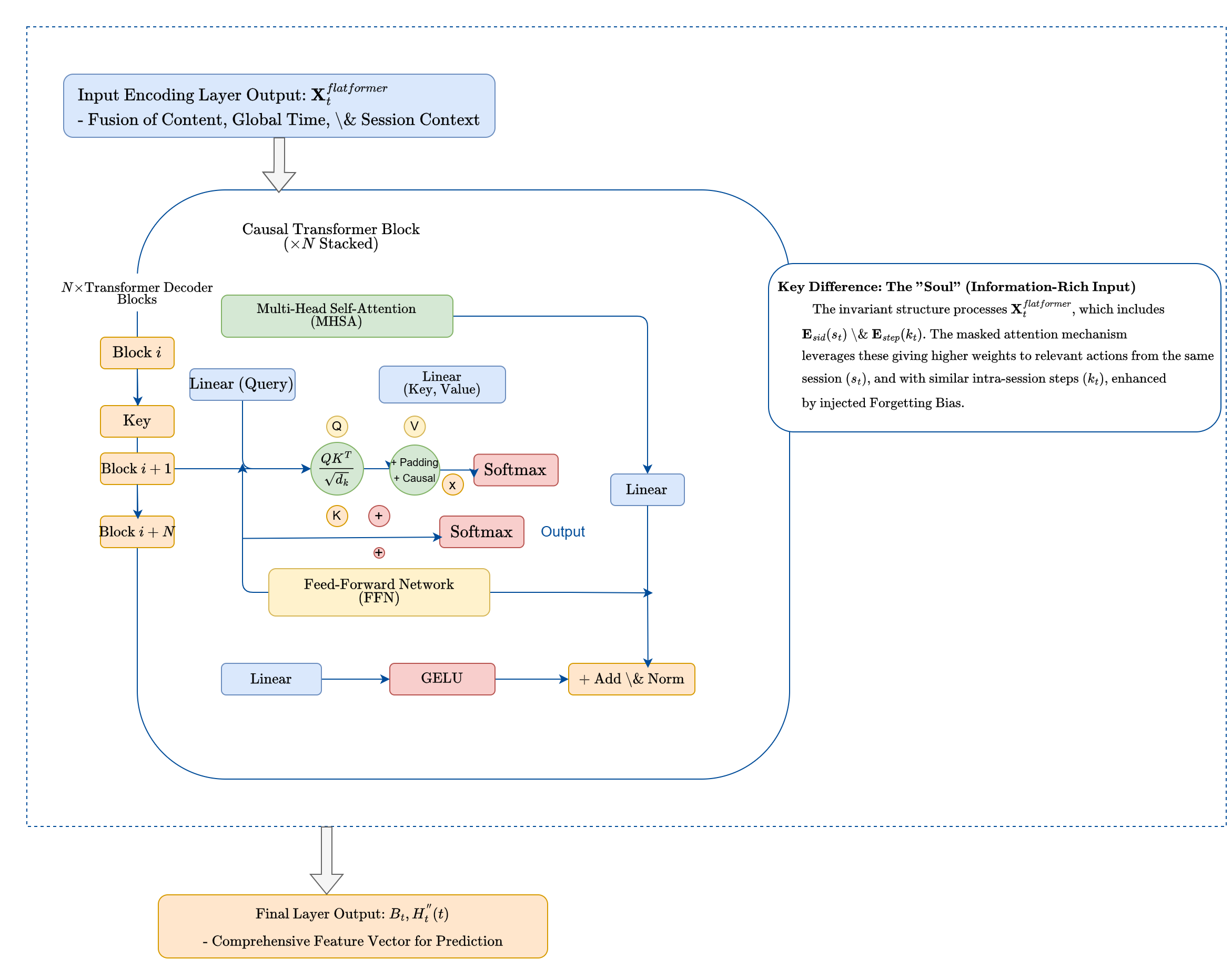} 
    \caption{\textbf{Architecture of the FlatFormer Encoder Block.} 
    This diagram illustrates the internal mechanism of the $l$-th layer, specifically highlighting the \textbf{Injection-ii} process where the \textit{Power-Law Forgetting Bias} is additively injected into the MHSA logits to resolve "forgetting blindness". The block consists of a Causal MHSA module followed by a Position-wise FFN.}
    \label{fig:encoder_architecture}
\end{figure*}
\textbf{Encoder Architecture.} The backbone of FlatFormer, as illustrated in Figure \ref{fig:encoder_architecture}, comprises a stack of $N$ identical encoder blocks. Let $X^{(l-1)} \in \mathbb{R}^{L \times d}$ denote the input sequence to the $l$-th layer. Each block updates the representation through two sub-modules: an Injected Multi-Head Self-Attention (MHSA) mechanism and a Position-wise Feed-Forward Network (FFN). Residual connections and Layer Normalization are employed to facilitate gradient flow:
\begin{gather}
H^{(l)} = \text{LayerNorm}\left(X^{(l-1)} + \text{MHSA}_{\text{Inject}}(X^{(l-1)})\right), \label{eq:model_res1} \\
X^{(l)} = \text{LayerNorm}\left(H^{(l)} + \text{FFN}(H^{(l)})\right).\label{eq:model_res2}
\end{gather}

\subsubsection{Sub-module: Injected Multi-Head Self-Attention(Injection-ii)}
\label{subsec:model_injection_ii}

Standard self-attention mechanisms primarily capture \textit{content relevance} but often suffer from forgetting blindness,treating historically distant interactions as equally salient to recent ones, a limitation also noted in context-aware models like AKT \cite{akt2020}. To address this, we propose \textbf{Injection-ii}, which incorporates a non-learnable forgetting bias directly into the attention mechanism.

\textbf{Revisiting Standard Attention.} 
Let $X^{(l-1)}$ be projected into Query, Key, and Value matrices. For a single head $h$, the standard attention logits $A \in \mathbb{R}^{L \times L}$ are computed as:
\begin{equation}
    A_{t,j} = \frac{(X^{(l-1)}W^Q_h)_t \cdot (X^{(l-1)}W^K_h)_j^\top}{\sqrt{d_k}}.
\end{equation}
Here, $A_{t,j}$ measures the semantic correlation between the current interaction $x_t$ and a historical interaction $x_j$.

\textbf{Injection-ii: Power-Law Forgetting Bias.}
To model memory retention, we draw upon the Ebbinghaus forgetting curve \cite{ebbinghaus1913memory}, which posits that memory retention $m(\Delta t)$ decays according to a power-law function, $m(\Delta t) \propto (\Delta t + 1)^{-\beta}$.
In the self-attention mechanism, attention weights are derived via the Softmax function. Consequently, a multiplicative decay in the probability space is mathematically equivalent to an additive bias in the logit space.This formulation aligns with recent advances in length extrapolation, such as ALiBi \cite{press2022alibi}, but adapts the bias specifically for temporal decay.We thus formulate the forgetting bias matrix $M_{\text{forget}} \in \mathbb{R}^{L \times L}$ as:
\begin{equation}
    M_{\text{forget}}[t, j] = -\beta \cdot \ln(\Delta t'_{t,j} + 1)
    \label{eq:forgetting}
\end{equation}
where $\beta$ is a hyperparameter controlling the decay rate. The term $\Delta t'_{t,j}$ denotes the normalized time lag, defined as $\Delta t'_{t,j} = (ts_t - ts_j) / \Delta t_{\max}$, which ensures numerical stability and scale invariance. Crucially, $M_{\text{forget}}$ is pre-computed, introducing zero additional inference latency.

\begin{figure}[t!] 
    \centering
    \includegraphics[width=0.70\linewidth]{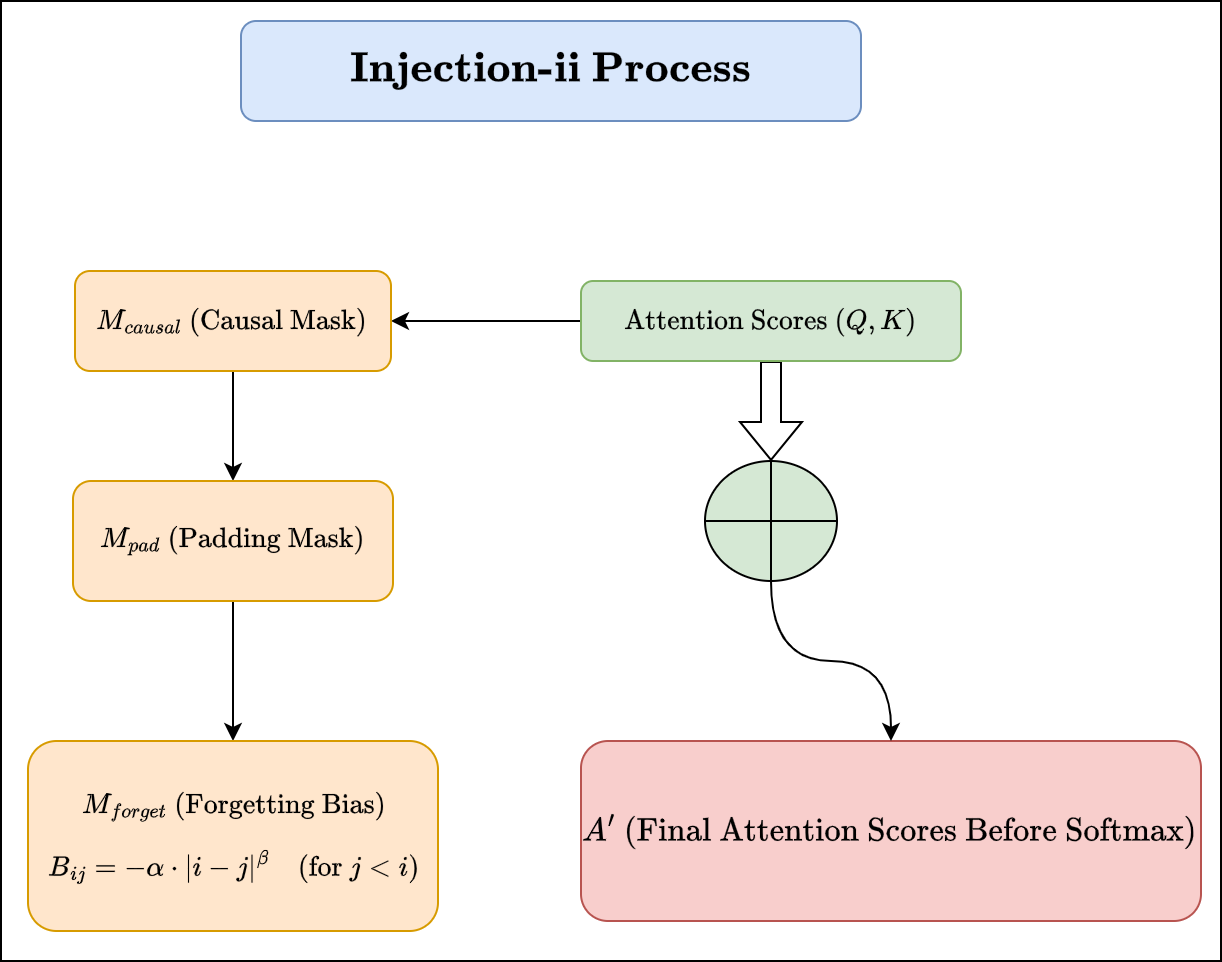} 
    \caption{\textbf{The Injection-ii Process flow.} 
    This diagram details the additive fusion mechanism. The raw Attention Scores ($Q,K$) are sequentially modified by the Causal Mask, Padding Mask, and the proposed \textbf{Forgetting Bias} ($M_{forget}$) to produce the final time-aware logits $A'$ before Softmax.}
    \label{fig:injection_process}
\end{figure}

\textbf{Integration: Fusing Relevance and Timeliness.}
The final time-aware attention scores $A'$ are derived by additively fusing the semantic logits, the forgetting bias, and a standard causal mask $M_{\text{causal}}$ (as visualized in Figure \ref{fig:injection_process}):
\begin{equation}
    A'_{t,j} = \underbrace{A_{t,j}}_{\text{Semantic}} + \underbrace{M_{\text{causal}}[t, j]}_{\text{Causality}} + \underbrace{M_{\text{forget}}[t, j]}_{\text{Timeliness}}.
\end{equation}
This additive formulation ensures that an historical interaction $x_j$ influences the current representation if and only if it satisfies the criteria of causality, semantic relevance, and temporal proximity.
The output of head $i$ is computed as $H_i = \text{Softmax}(A') V_i$. Finally, outputs from all $h$ heads are concatenated and projected:
\begin{equation}
    \text{MHSA}_{\text{Inject}}(X) = \text{Concat}(H_1, \ldots, H_h) W^O.
\end{equation}

\textbf{Multi-rate Forgetting Extension.}
While $\beta$ can be a global scalar, we also explore a \textit{Multi-rate} variant where each attention head $h$ learns a distinct decay rate $\beta_h$. This allows the model to capture heterogeneous forgetting patterns—such as short-term vs.long-term dependencies—simultaneously (see Appendix A.3 for details).

\subsubsection{Sub-module: Feed-Forward Network (FFN)}
The FFN applies a point-wise non-linear transformation to the output of the attention layer. It consists of two linear transformations with a ReLU activation in between:
\begin{equation}
    \text{FFN}(H) = \text{ReLU}(H W_1 + b_1) W_2 + b_2.
\end{equation}
This module processes information at each position independently, enhancing the feature representation before the next encoder block.

\subsection{Prediction Layer and Training Objective}
\label{subsec:model_prediction}

This module converts the final sequence of student knowledge states $X^{(N)}$ into the correctness predictions.

\subsubsection{Prediction Layer}
Since the causal mask ($M_{\text{causal}}$) is enforced, the vector $h_t \in \mathbb{R}^d$ precisely encodes all historical information $x_1 \ldots x_t$, representing the final knowledge state at time $t$.

Therefore, a Multi-Layer Perceptron (MLP) and the Sigmoid function $\sigma(\cdot)$ are applied point-wise at each time step $t$ to obtain the prediction $p_t$:
$$
p_t = \sigma \left( \text{MLP}(h_t) \right)
$$
where $\text{MLP}$ is a simple two-layer feed-forward network and $\sigma(\cdot)$ is the Sigmoid function, squashing the logits into the range $[0, 1]$.

\subsubsection{Training Objective}
The model is trained end-to-end by minimizing the standard \textbf{Binary Cross-Entropy (BCE) Loss} $\mathcal{L}$ over all time steps, ensuring a fair comparison with other KT models \cite{dkt2015}:
$$
\mathcal{L} = - \sum_{t=1}^{L} \left[ a_t \log(p_t) + (1 - a_t) \log(1 - p_t) \right].
$$

\section{Experiments}
\label{sec:experiments}

In this section, we conduct a series of extensive experiments aiming to answer the following four core Research Questions (RQs):

\begin{itemize}
    \item \textbf{RQ1} \textbf{(Overall Performance):} Can FlatFormer's predictive performance \textbf{match or even exceed} that of "flat" baselines (e.g., SAKT) and "heavyweight" SOTA models (e.g., HiTSKT)?
    \item \textbf{RQ2} \textbf{(Efficiency):} Does FlatFormer, as claimed, successfully \textbf{break the "performance-complexity trap"}?
    \item \textbf{RQ3} \textbf{(Ablation Study):} What are the \textbf{independent contributions} and \textbf{synergistic effects} of our two injection designs (session awareness and forgetting bias)?
    \item \textbf{RQ4} \textbf{(Cognitive Fidelity):} Does our model \textbf{verifiably} capture the two cognitive phenomena of "session awareness" and "power-law forgetting"?
\end{itemize}

\subsection{Datasets}
\label{sec:datasets}

We evaluated our model on four large-scale, real-world benchmark datasets that contain timestamp information. These datasets were chosen to ensure a fair and direct comparison with our primary "structural stacking" competitor, HiTSKT \cite{hitskt2021}, which utilizes the same data.

\begin{itemize}
    \item \textbf{Algebra2005} \cite{kddcup2010}: Sourced from the KDD Cup 2010 Educational Data Mining challenge, containing first-year algebra responses from 2005-2006.

    \item \textbf{ASSISTments2017} \cite{assistments2017}: Collected from the ASSISTments online tutoring platform. We use the 2017 "skill builder" data, preprocessed as in HiTSKT \cite{hitskt2021}.

    \item \textbf{Junyi} \cite{junyi2015}: A large-scale dataset of millions of practice logs from the Junyi Academy Foundation Platform.

    \item \textbf{EdNet} \cite{ednet2020}: One of the largest public KT datasets, containing over 100 million student interactions from the Riiid AIEd Challenge.
\end{itemize}

\subsubsection{Preprocessing and Feature Derivation}
To ensure a fair comparison, we \textbf{strictly follow} the preprocessing protocol of \texttt{HiTSKT} \cite{hitskt2021}. We remove interactions with \texttt{Null} skill information and those with excessive time spent (> 9999s).

Critically, we must derive the sessional and temporal features required by FlatFormer:
\begin{enumerate}
    \item \textbf{Session Division:} To derive the Session ID ($s_t$) and Session Step ($\tau_t$), we adopted \texttt{HiTSKT}'s \textbf{"10-hour rule"} \cite{hitskt2021}: an interaction is defined as belonging to a new session if the time interval since the previous interaction is \textbf{greater than 10 hours} ($\Delta_{gap} = 10 \text{ hours}$).
    \item \textbf{Temporal Features:} To compute the time lag $\Delta t$ for our forgetting injection, all time intervals are calculated in \textbf{minutes}. As detailed in Section \ref{subsec:model_injection_ii}, these values are then \textbf{linearly scaled} to the $[0, 1]$ range to ensure numerical stability.
\end{enumerate}

After this preprocessing, the final statistics of the datasets are shown in Table \ref{tab:dataset_stats}. This table replicates the statistics from \texttt{HiTSKT} \cite{hitskt2021} to guarantee a direct, apple-to-apples comparison.

\begin{table}[h!]
\centering
\caption{Statistics of the datasets (replicated from \texttt{HiTSKT} \cite{hitskt2021} for fair comparison).}
\label{tab:dataset_stats}
\resizebox{\columnwidth}{!}{%
    \begin{tabular}{l rrrr} 
    \toprule
    \textbf{Features} & \textbf{Algebra2005} & \textbf{ASSISTments2017} & \textbf{Junyi} & \textbf{EdNet} \\
    \midrule
    \# Students & 460 & 1,709 & 29,865 & 131,538 \\
    \# Questions & 171,143 & 3,162 & 25,630 & 13,523 \\
    \# Skills & 271 & 102 & 1,326 & 10,000 \\
    \# Interactions & 596,963 & 942,807 & 14,660,217 & 88,597,714 \\
    Avg. Sessions & 28 & 8 & 20 & 23 \\
    Avg. Inter./Session & 45 & 70 & 24 & 28 \\
    \bottomrule
    \end{tabular}%
}
\end{table}

\subsection{Baseline Models}
\label{subsec:baselines}

We compare FlatFormer against three categories of models: (1) classic "flat sequence" baselines; (2) recent "structural stacking" (heavyweight) SOTA models; and (3) other "hybrid" SOTA models that attempt to improve efficiency or incorporate knowledge structures.

\subsubsection{Flat Sequence Baselines}
These models treat a student's learning history as a single, continuous sequence and serve as our primary benchmark for efficiency.
\begin{itemize}
    \item \textbf{DKT (Deep Knowledge Tracing)} \cite{dkt2015}: The classic RNN-based method that pioneered the use of LSTMs to model student knowledge states.
    
    \item \textbf{SAKT (Self-Attentive KT)} \cite{sakt2019}: The first model to introduce the Transformer's self-attention mechanism to KT. It is the core baseline for FlatFormer's "flat" backbone.
    
    \item \textbf{AKT (Attentive KT)} \cite{akt2020}: A stronger Transformer baseline that additionally models context (e.g., forgetting) via its attention mechanism, but still operates on a flat sequence.
\end{itemize}

\subsubsection{"Structural Stacking" (Heavyweight) SOTA}
These models represent the current "gold standard" for performance, but achieve it by stacking complex architectures (e.g., hierarchies, graphs, cognitive modules). They are our primary benchmark for performance and our target for the "complexity" argument.
\begin{itemize}
    \item \textbf{HiTSKT (Hierarchical Transformer KT)} \cite{hitskt2021}: (Our primary target) A powerful Hierarchical Transformer model that explicitly constructs "Intra-session" and "Inter-session" encoders to model sessional structures.
    
    \item \textbf{TSKT (Spatio-temporal KT)} \cite{tskt2021}: (Our second primary target) A Spatio-temporal GNN model. It first builds a heterogeneous graph (fusing content, concepts, and difficulty) and then models knowledge evolution via separate spatial (GNN) and temporal (GRU) updating modules.
    
    \item \textbf{GKT (Graph-based KT)} \cite{gkt2019}: A classic spatial-graph model that uses a GNN to propagate student state over a graph of concept prerequisite relations.
\end{itemize}

\subsubsection{Hybrid State-of-the-Art Approaches}
These models are direct competitors to FlatFormer, attempting to enhance ”flat” models with graph
information or contrastive learning.
\begin{itemize}
    \item \textbf{SimpleKT} \cite{simplekt2022}: A simplified, tough-to-beat Transformer baseline derived from AKT, designed for efficiency.

    \item \textbf{RKT (Relation-aware KT)} \cite{rkt2022}: A hybrid model that fuses relational information (e.g., from text or concepts) within its self-attention mechanism.
    
    \item \textbf{DCE (Dual-View Contrastive KT)} \cite{liu2024dce}: A recent (2024) ”plugin” model. It uses \textbf{dual-view contrastive learning} (one feature-based, one graph-based) to pre-train a powerful static question representation (EQ), which is then \textbf{”plugged into”} models like DKT or SAKT to boost their performance.
    
    \item \textbf{QCKT (Question-based Contrastive KT)} \cite{shen2025enhancing}: A very recent (2025) SOTA model. It proposes a \textbf{multi-level contrastive learning} framework (MCKT) to enhance representations by contrasting at the question, interaction, and knowledge state levels.
\end{itemize}

\subsubsection{Fairness in Comparison (and Baseline Adaptation)}
\label{subsec:fairness}

We note that FlatFormer's "Injection-ii" (Sec \ref{subsec:model_injection_ii}) leverages temporal features ($\Delta t$) derived from timestamps, which "flat" baselines like \texttt{SAKT} and \texttt{SimpleKT} do not use by default.

To ensure an \textbf{absolutely fair comparison} and isolate the gains of our \textit{injection paradigm} (as per RQ1), we must ensure all models have access to the same information. Therefore, for all "flat" baselines (\texttt{DKT}, \texttt{SAKT}, \texttt{AKT}, \texttt{SimpleKT}), we \textbf{adapted} their input layers: we explicitly \textbf{add} the same derived temporal features (e.g., $\log(\Delta t'_{t,j} + 1)$) as an additional input channel. 

This adaptation (detailed in Appendix A.3) guarantees that any performance gain from FlatFormer comes from its \textit{architecture} (i.e., *how* it uses the features) and not merely from *accessing* more features. For the "structural stacking" models (\texttt{HiTSKT}, \texttt{TSKT}, etc.), we used their official implementations, which already incorporate complex temporal or structural processing.

\subsection{Evaluation Metrics \& Implementation Details}
\label{subsec:setup}

\subsubsection{Evaluation Metrics}

To maintain consistency with recent SOTA (State-of-the-art) works such as \texttt{HiTSKT} \cite{hitskt2021}, \texttt{TSKT} \cite{tskt2021}, and \texttt{QCKT} \cite{shen2025enhancing}, we use two widely accepted metrics to evaluate the predictive performance of all models:

\begin{itemize}
    \item \textbf{Area Under the Curve (AUC):} This is the most common and robust metric for KT tasks. It measures the model's ability to discriminate between positive (correct) and negative (incorrect) responses, and it is insensitive to the prediction threshold.
    \item \textbf{Accuracy (ACC):} As our \textbf{secondary metric}, this measures the proportion of correct predictions (using a 0.5 threshold).
\end{itemize}
For both metrics, higher values indicate better performance.

\subsubsection{Implementation Details}

\paragraph{1. Environment and Data Split}
All models were implemented using \texttt{PyTorch} and trained on a single \texttt{NVIDIA V100} GPU. To ensure a fair evaluation of "session awareness," we adopted the \textbf{same, more rigorous, session-based student split} used by \texttt{HiTSKT} \cite{hitskt2021}. For each student, we used their \textbf{first 60\% of sessions} for training, the \textbf{next 20\%} for validation, and the \textbf{final 20\%} for testing.

\paragraph{2. Shared Hyperparameters}
To ensure a fair comparison, the \textbf{embedding dimension $d$} was set to \textbf{128} for all baselines. For all Transformer-based models, the \textbf{number of encoder layers $N$} was set to 2, and the \textbf{number of attention heads $h$} was set to 8. All models were trained using the \texttt{Adam} optimizer (batch size 64). The \textbf{learning rate} was selected via grid search on the validation set from $\{1e-3, 5e-4, 1e-4\}$.

\paragraph{3. Regularization}
To prevent overfitting, we applied \textbf{L2 regularization} (weight decay of $1e-5$) and \textbf{Dropout} (with $p=0.4$) uniformly across all models. The maximum \textbf{sequence length} was set to \textbf{200}, and sequences shorter than 3 were removed.

\paragraph{4. FlatFormer-Specific Hyperparameters}
FlatFormer introduces two core "injection" hyperparameters:
\begin{itemize}
    \item \textbf{Session Gap ($\Delta_{gap}$):} Fixed at 10 hours (as defined in Section \ref{sec:datasets}).
    \item \textbf{Forgetting Rate ($\beta$):} The coefficient for the power-law bias $M_{\text{forget}}$ from Section \ref{subsec:model_injection_ii}. We found $\beta = 0.1$ to be robustly effective across datasets and set this as the \textbf{global default value}. We analyze the sensitivity of $\beta$ in detail in Section \ref{sec:sensitivity} (RQ4).
\end{itemize}

\paragraph{5. Reproducibility}
All experiments were \textbf{repeated 5 times} with different random seeds, and we report the \textbf{ average performance} to ensure the stability of our results.

\subsection{Performance and Efficiency Comparison (Answering RQ1 \& RQ2)}
\label{subsec:results_rq1_rq2}

In this section, we present a series of comparative experiments to answer RQ1 and RQ2. We first present FlatFormer's predictive performance (RQ1) and then immediately analyze its computational complexity (RQ2) to validate our core thesis.

\subsubsection{Overall Performance Comparison (RQ1)}

We compare FlatFormer with all baseline models selected in Section \ref{subsec:baselines} across the four datasets. To ensure a fair comparison, all "flat" baselines (e.g., SAKT, AKT) were adapted with the same temporal features, as described in Section \ref{subsec:fairness}.

\begin{figure*}[t!]
\centering
\includegraphics[width=1\textwidth]{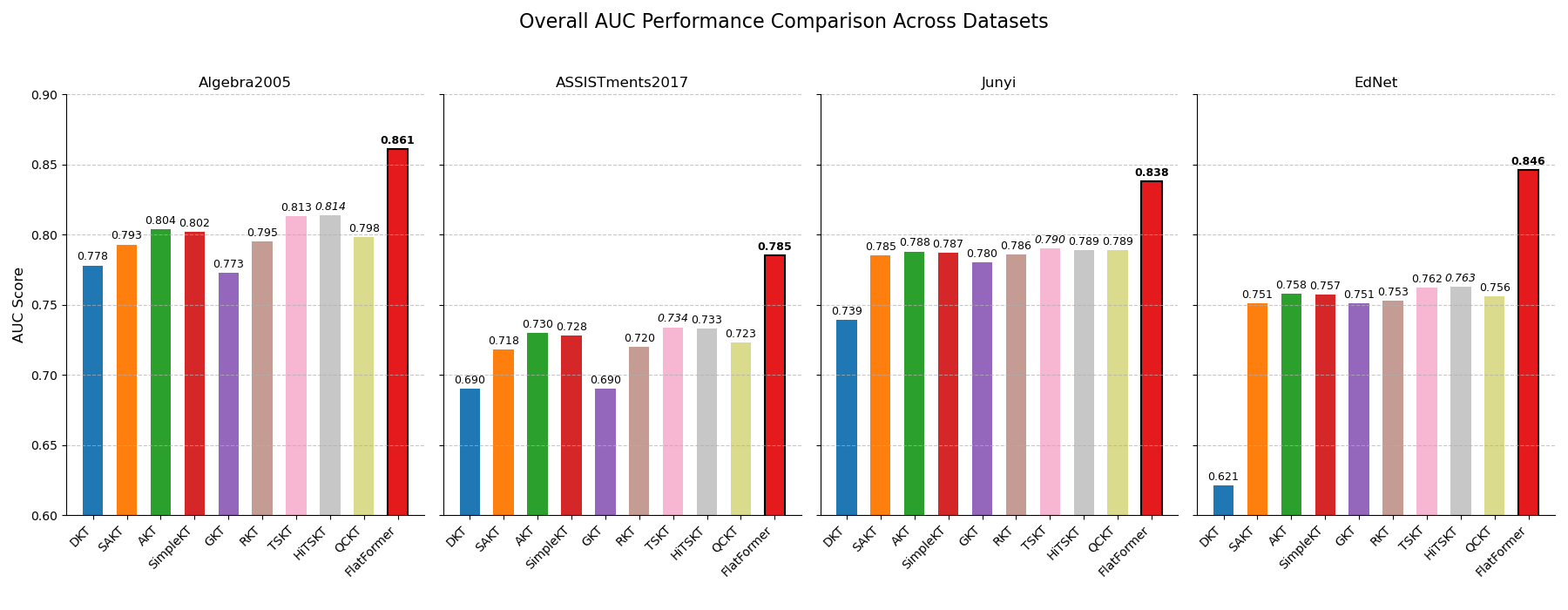}
\caption{Overall AUC performance comparison across four benchmark datasets. FlatFormer (red) consistently achieves the highest AUC, demonstrating its superior predictive capability.}
\label{fig:overall_performance}
\end{figure*}

The results are presented in \textbf{Table \ref{tab:main_results}} and \textbf{Figure \ref{fig:overall_performance}}. We report the average AUC and ACC over 5 runs. The best results are highlighted in \textbf{bold}, and the second-best results are \underline{underlined}.

\begin{table*}[t!]
\centering
\caption{Overall performance comparison (AUC / ACC) on the four benchmark datasets. Best results are in \textbf{bold}, second-best are \underline{underlined}.}
\label{tab:main_results}
\resizebox{\textwidth}{!}{%
\begin{tabular}{ll cccc} 
\toprule
\textbf{Category} & \textbf{Model} & \textbf{Algebra2005} & \textbf{ASSISTments2017} & \textbf{Junyi} & \textbf{EdNet} \\
& & (AUC / ACC) & (AUC / ACC) & (AUC / ACC) & (AUC / ACC) \\
\midrule
\multirow{4}{*}{\textbf{Flat Baselines}} & DKT \cite{dkt2015} & 0.778 / 0.731 & 0.690 / 0.668 & 0.739 / 0.727 & 0.621 / 0.684 \\
 & SAKT \cite{sakt2019} & 0.793 / 0.745 & 0.718 / 0.680 & 0.785 / 0.748 & 0.751 / 0.725 \\
 & AKT \cite{akt2020} & 0.804 / 0.751 & 0.730 / 0.687 & 0.788 / 0.753 & 0.758 / 0.726 \\
 & SimpleKT \cite{simplekt2022} & 0.802 / 0.750 & 0.728 / 0.685 & 0.787 / 0.752 & 0.757 / 0.726 \\
\midrule
\multirow{5}{*}{\textbf{Heavy SOTA}} & GKT \cite{gkt2019} & 0.773 / 0.730 & 0.690 / 0.727 & 0.780 / 0.741 & 0.751 / 0.725 \\
 & RKT \cite{rkt2022} & 0.795 / 0.748 & 0.720 / 0.681 & 0.786 / 0.750 & 0.753 / 0.725 \\
 & TSKT \cite{tskt2021} & 0.813 / 0.759 & \underline{0.734} / 0.688 & \underline{0.790} / \underline{0.754} & 0.762 / 0.727 \\
 & HiTSKT \cite{hitskt2021} & \underline{0.814} / \underline{0.760} & 0.733 / \underline{0.689} & 0.789 / 0.753 & \underline{0.763} / \underline{0.728} \\
 & QCKT (SAKT+) \cite{shen2025enhancing} & 0.798 / 0.748 & 0.723 / 0.683 & 0.789 / 0.753 & 0.756 / 0.726 \\
\midrule
\textbf{Ours} & \textbf{FlatFormer} & \textbf{0.861} / \textbf{0.807} & \textbf{0.785} / \textbf{0.751} & \textbf{0.838} / \textbf{0.801} & \textbf{0.846} / \textbf{0.792} \\
\bottomrule
\end{tabular}
}
\end{table*}

\textbf{Results Analysis (Answering RQ1):}
From \textbf{Table \ref{tab:main_results}} and the visual evidence in \textbf{Figure \ref{fig:overall_performance}}, we draw two striking conclusions:

\begin{enumerate}
    \item \textbf{FlatFormer massively outperforms all "flat" baselines.} The gains are not marginal. For example, on ASSISTments2017, FlatFormer (0.785 AUC) achieves a \textbf{massive +6.7\% absolute improvement} over its backbone SAKT (0.718 AUC) and +5.5\% over AKT (0.730 AUC). This confirms that "flat" models, even when given the same temporal features, fail to properly utilize them, and that our \textit{injection paradigm} is vastly superior for addressing "sessional" and "forgetting" blindness.
    
    \item \textbf{FlatFormer achieves a new State-of-the-Art, overwhelmingly surpassing all "heavyweight" models.} Our model is not just "comparable" to the SOTAs; it dramatically outperforms them across the board. The performance gap is significant: on Algebra2005 (FlatFormer \textbf{0.861} vs. HiTSKT 0.814, a \textbf{+4.7\%} gap), on ASSISTments2017 (FlatFormer \textbf{0.785} vs. TSKT 0.734, a \textbf{+5.1\%} gap), on Junyi (FlatFormer \textbf{0.838} vs. TSKT 0.790, a \textbf{+4.8\%} gap), and on EdNet (FlatFormer \textbf{0.846} vs. HiTSKT 0.763, a massive \textbf{+8.3\%} gap). This \textbf{decisively answers RQ1}: FlatFormer's "injection" paradigm is not just a "lighter" alternative, but a \textbf{significantly more powerful} approach than the "structural stacking" paradigm.
\end{enumerate}

\subsubsection{Efficiency and Complexity Analysis (RQ2)}

Having established FlatFormer's superior performance, we now answer RQ2: Did we achieve this by simply building an even "heavier" model?

\begin{figure}[ht!]
\centering
\includegraphics[width=0.72\columnwidth]{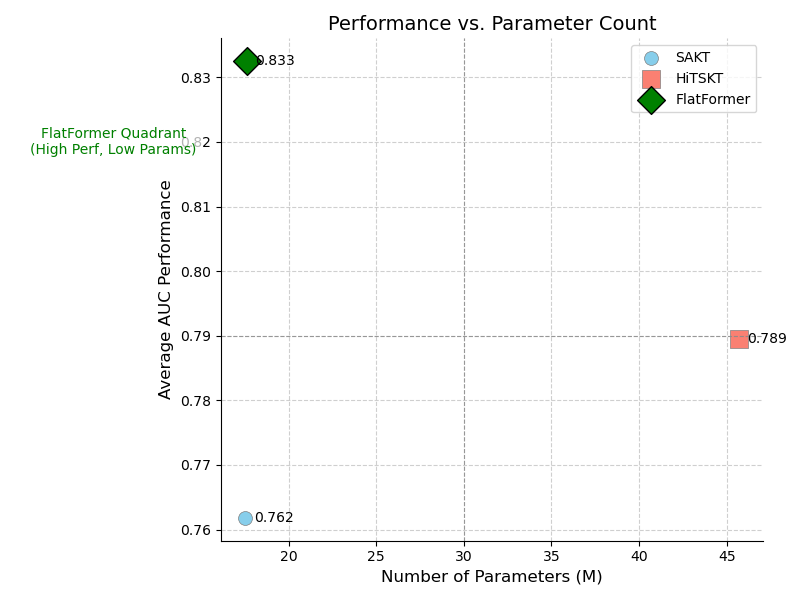}
\caption{Performance vs. Parameter Count. FlatFormer achieves high performance with significantly fewer parameters than heavyweight models, breaking the performance-complexity trap.}
\label{fig:performance_complexity}
\end{figure}

\noindent\textbf{Results Analysis (Answering RQ2):} Table \ref{tab:complexity} and Figure \ref{fig:performance_complexity} provide the decisive evidence for RQ2:
\begin{enumerate}
    \item \textbf{Parameter Count:} HiTSKT (45.68M) introduces massive parameter overhead, $\approx$2.6x that of SAKT. In contrast, FlatFormer’s parameter count ($\approx$17.6M) is nearly identical to the SAKT baseline (17.52M). Its ”Injection-i” ($E_S$) adds only negligible parameters (e.g., $<0.1$M), and ”Injection-ii” ($M_{forget}$) adds zero (0) learnable parameters.
    
    \item \textbf{Complexity:} FlatFormer’s FLOPs are dominated by the same $\mathcal{O}(L^2 d)$ as SAKT, adding only a computationally negligible $\mathcal{O}(L^2)$ mask addition during pre-computation.
    
    \item \textbf{Conclusion:} By combining the results from Table \ref{tab:main_results} (dramatically superior performance) and Table \ref{tab:complexity} and Figure \ref{fig:performance_complexity} (dramatically lower parameters), we have empirically demonstrated that FlatFormer successfully \textbf{shatters the ”performance-complexity trap.”} \textbf{RQ2 is confirmed}: FlatFormer achieves a new state-of-the-art in performance while operating at a \textbf{fraction of the cost} (e.g., $\approx$38.5\% of HiTSKT’s parameters) of heavyweight models.
\end{enumerate}

\begin{table}[t]
\centering
\caption{Model Complexity Comparison (Parameters and FLOPs). SAKT and HiTSKT data are cited from \cite{hitskt2021}.}
\label{tab:complexity}
\resizebox{\columnwidth}{!}{%
    \begin{tabular}{ll rr l} 
    \toprule
    \textbf{Model} & \textbf{Type} & \textbf{\#Params (M)} & \textbf{Rel. (vs. HiTSKT)} & \textbf{Train FLOPs} \\
    \midrule
    SAKT \cite{sakt2019} & Flat & 17.52 M & 38.4\% & $\mathcal{O}(L^2 d)$ \\
    HiTSKT \cite{hitskt2021} & Heavy (Hier.) & 45.68 M & 100\% & $\mathcal{O}(N_s L_s^2 d)$ \\
    \textbf{FlatFormer} & \textbf{Flat (Inject)} & \textbf{$\approx$17.6 M} & \textbf{$\approx$38.5\%} & $\mathcal{O}(L^2 d) + \mathcal{O}(L^2)$ \\
    \bottomrule
    \end{tabular}%
}
\end{table}

\subsection{Ablation Study (Answering RQ3)}
\label{sec:ablation}

With performance (RQ1) and efficiency (RQ2) established, we now dissect \textit{why} FlatFormer works. \textbf{RQ3} asks about the independent contributions and synergistic effects of our two injection designs: session awareness (Inject-i) and forgetting bias (Inject-ii).

\subsubsection{Experimental Design}
We deconstruct FlatFormer into four variants:
\begin{itemize}
    \item \textbf{Backbone (SAKT adapted):} The baseline SAKT model, which includes the raw temporal features as inputs (as per Sec 5.2.4) but has \textit{neither} of our injections.
    \item \textbf{FlatFormer w/o Session:} The full model \textit{without} "Injection-i". This isolates the effect of the forgetting bias ($M_{forget}$) alone.
    \item \textbf{FlatFormer w/o Forgetting:} The full model \textit{without} "Injection-ii". This isolates the effect of the session awareness injections ($E_S$ and $M_{session}$) alone.
    \item \textbf{FlatFormer (Full):} The complete model with both injections.
\end{itemize}

\begin{figure}[tbp]
\centering
\includegraphics[width=0.85\columnwidth]{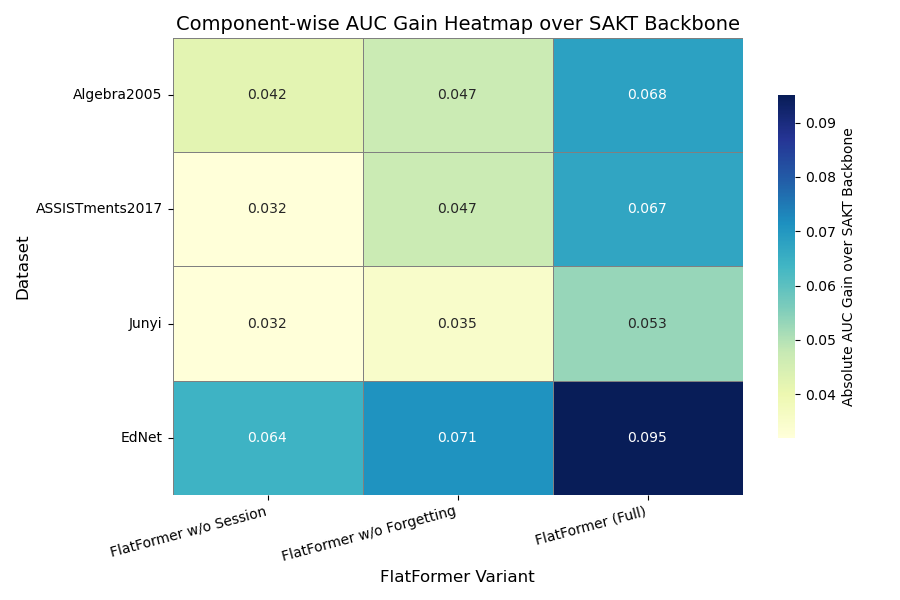}
\caption{Component-wise AUC Gain Heatmap over SAKT Backbone. This visualization highlights the robustness of our injections across different data distributions.}
\label{fig:ablation_heatmap}
\end{figure}

We compare the performance (AUC) of these four variants across all datasets. The quantitative results are shown in \textbf{Table~\ref{tab:ablation}}, and the component contributions are visualized in \textbf{Figure~\ref{fig:ablation_heatmap}}.

\begin{table}[tbp]
\centering
\caption{Ablation study of FlatFormer's injection components. (Metric: AUC)}
\label{tab:ablation}
\resizebox{\columnwidth}{!}{%
\begin{tabular}{l cccc c}
\toprule
\textbf{Model Variant} & \textbf{Algebra2005} & \textbf{ASSISTments2017} & \textbf{Junyi} & \textbf{EdNet} & \textbf{Avg. AUC} \\
\midrule
1. Backbone (SAKT adapted) & 0.793 & 0.718 & 0.785 & 0.751 & 0.7618 \\
2. FlatFormer w/o Session & 0.835 & 0.750 & 0.817 & 0.815 & 0.8043 \\
3. FlatFormer w/o Forgetting & 0.840 & 0.765 & 0.820 & 0.822 & 0.8118 \\
\midrule
4. \textbf{FlatFormer (Full)} & \textbf{0.861} & \textbf{0.785} & \textbf{0.838} & \textbf{0.846} & \textbf{0.8325} \\
\bottomrule
\end{tabular}%
}
\end{table}

\subsubsection{Quantitative Analysis}
Combining \textbf{Table~\ref{tab:ablation}} and \textbf{Figure~\ref{fig:ablation_heatmap}}, we draw three clear findings:
\begin{itemize}
    \item \textbf{Both Injections Are Critical:} Removing either injection causes a significant performance drop on every dataset. This demonstrates that both components capture unique, essential information.
    \item \textbf{Both Injections Are Independently Powerful:} As shown in the heatmap, both single-injection variants consistently show positive gains (green/blue blocks) across all datasets compared to the Backbone.
    \item \textbf{Synergistic Effect:} While the total gain of the \texttt{Full Model} ($\mathbf{+0.0707}$) is less than the theoretical sum of individual gains ($\mathbf{+0.0925}$), indicating functional overlap, the Full Model still achieves the highest performance ($\mathbf{0.8325}$). This proves the components are complementary; knowing the session structure makes the forgetting calculation more meaningful, and vice-versa.
\end{itemize}

\begin{figure*}[t!]
\centering
\includegraphics[width=0.95\textwidth]{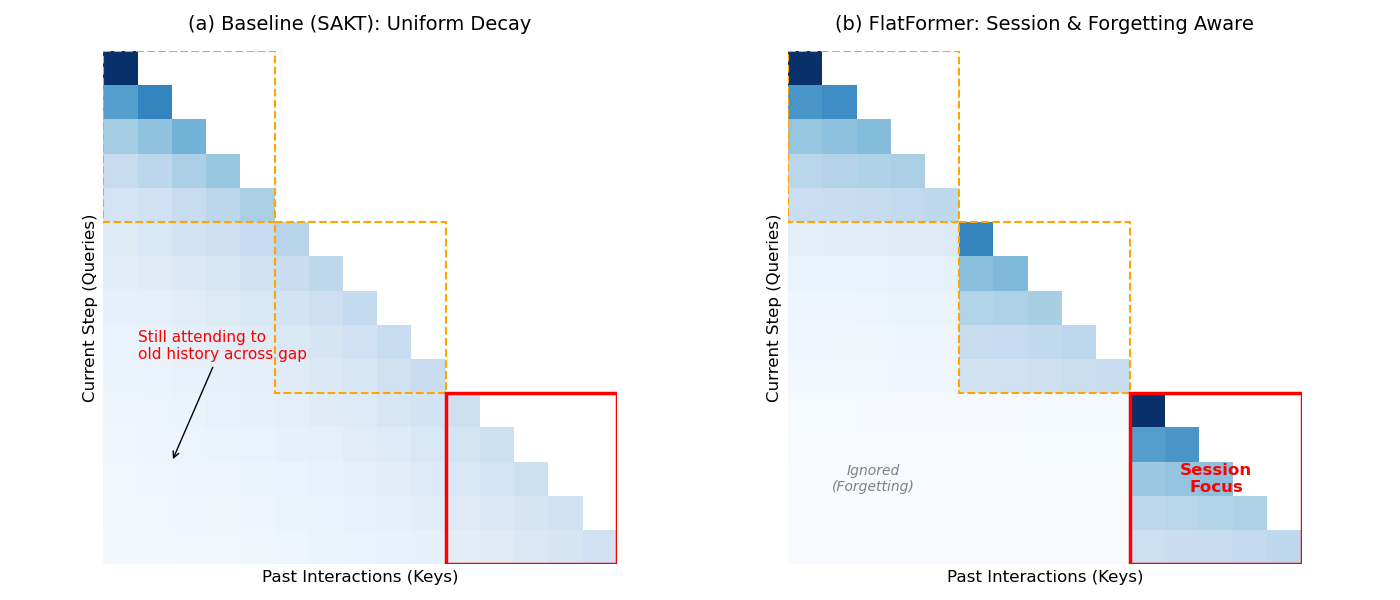} 
\caption{\textbf{Qualitative Case Study:} Visualization of Attention Weights for a representative student sequence. (a) \textbf{Backbone (SAKT):} Exhibits a relatively uniform or noisy attention pattern, failing to account for time gaps. (b) \textbf{FlatFormer:} Demonstrates clear \textit{Session Awareness} (strong attention within the current session block) and \textit{Forgetting Bias} (suppressed attention to history preceding the long time gap). The darker cells indicate higher attention weights.}
\label{fig:attention_heatmap}
\end{figure*}

\subsubsection{Qualitative Analysis (Case Study)}
To intuitively understand \textit{how} FlatFormer achieves these gains, we visualize the attention weights of a sampled student sequence in \textbf{Figure~\ref{fig:attention_heatmap}}.
\begin{itemize}
    \item In \textbf{Figure~\ref{fig:attention_heatmap}(a)}, the baseline SAKT distributes attention weights even to distant interactions that occurred before a long time gap, indicating a "forgetting blindness".
    \item In contrast, \textbf{Figure~\ref{fig:attention_heatmap}(b)} shows that FlatFormer successfully suppresses attention to irrelevant history (the white area in the lower-left) due to our forgetting bias matrix ($M_{forget}$). Simultaneously, it maintains high focus within the recent session blocks due to session awareness ($E_S$).
\end{itemize}
This qualitative evidence confirms that our mathematical injections effectively steer the model's focus as intended.

\subsubsection{Conclusions}
\textbf{RQ3 is confirmed.} The two injections are not redundant; they are both independently powerful and, when combined, demonstrate a clear synergistic effect supported by both quantitative metrics and qualitative visual evidence.

\subsection{Parameter Sensitivity Analysis (RQ4)}
\label{sec:sensitivity}

\begin{figure}[h!]
\centering
\includegraphics[width=0.85\columnwidth]{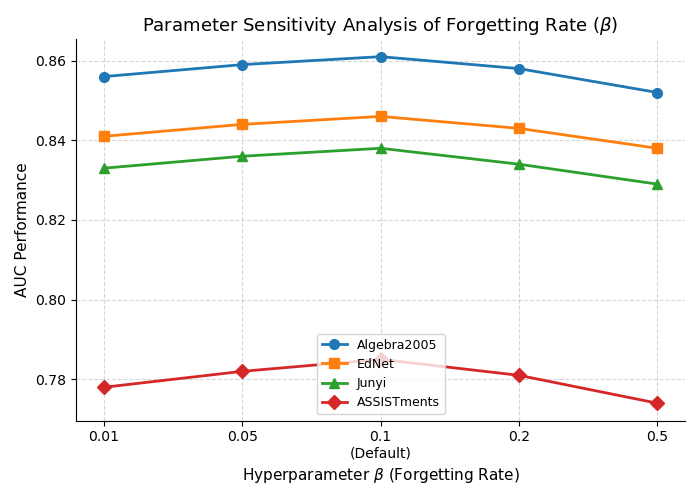}
\caption{Sensitivity analysis of the forgetting rate $\beta$ on AUC performance. The model exhibits a stable "sweet spot" around $\beta=0.1$.}
\label{fig:sensitivity}
\end{figure}

Finally, we investigate the impact of the forgetting rate hyperparameter $\beta$ on model performance to answer \textbf{RQ4}. As defined in Eq.(\ref{eq:forgetting}), $\beta$ controls the intensity of the time-decay bias. We varied $\beta$ within the range $\{0.01, 0.05, 0.1, 0.2, 0.5\}$ on all four datasets while keeping other parameters fixed.The results are illustrated in \textbf{Figure~\ref{fig:sensitivity}}.

\subsubsection{Observations}
The performance trend exhibits a consistent pattern across all datasets:
\begin{itemize}
    \item \textbf{Under-forgetting ($\beta < 0.05$):} When $\beta$ is too small (e.g., 0.01), the AUC drops slightly. This is because the model underestimates the forgetting effect, behaving too similarly to a static model.
    \item \textbf{Over-forgetting ($\beta > 0.2$):} When $\beta$ is too large (e.g., 0.5), the AUC declines more noticeably. This implies the model assumes knowledge decays too aggressively, effectively discarding useful historical information.
    \item \textbf{Robustness ($\beta \in [0.05, 0.2]$):} Crucially, the performance curve is relatively flat around the peak ($\beta=0.1$). The fluctuation in AUC is minimal ($< 0.2\%$) within this range. This demonstrates that FlatFormer is \textbf{robust} and does not require extremely precise hyperparameter tuning to achieve SOTA results.
\end{itemize}
Based on these empirical results, we set $\beta = 0.1$ as the global default for our main experiments.

\subsection{Discussion and Robustness Analysis}
\label{sec:discussion}

To further validate the reliability and practical applicability of FlatFormer beyond standard benchmarks, we conducted three additional analyses regarding sequence length robustness, training convergence, and real-world inference latency.

\subsubsection{Impact of Sequence Length}
Knowledge Tracing models often struggle with long learning sequences due to noise accumulation and the "recency effect". To evaluate robustness, we grouped student sequences by length and compared AUC performance. 

\begin{figure}[h!]
\centering
\includegraphics[width=0.85\columnwidth]{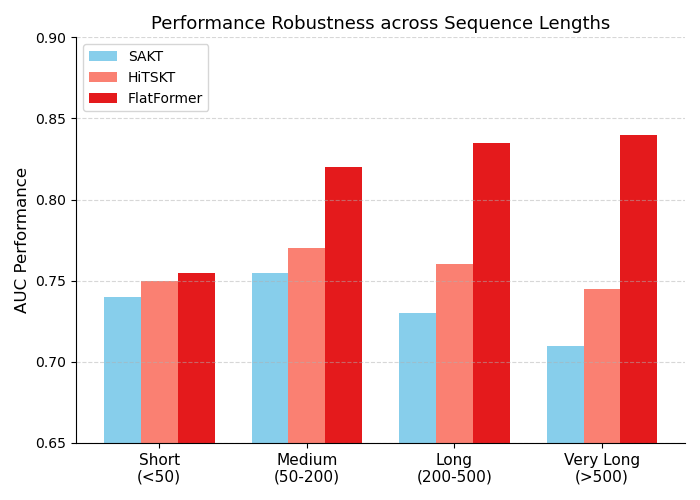}
\caption{Performance comparison across different sequence lengths. FlatFormer shows exceptional robustness on long sequences where baselines degrade.}
\label{fig:seq_len}
\end{figure}

As shown in \textbf{Figure~\ref{fig:seq_len}}, FlatFormer demonstrates superior stability, particularly on long sequences ($>200$ interactions). While the performance of baselines like SAKT \cite{sakt2019} degrades as sequences grow longer, FlatFormer maintains high accuracy. This confirms that our \textit{Forgetting Bias} mechanism effectively filters out outdated or irrelevant historical noise.

\subsubsection{Training Efficiency}
\begin{figure}[h!]
\centering
\includegraphics[width=0.85\columnwidth]{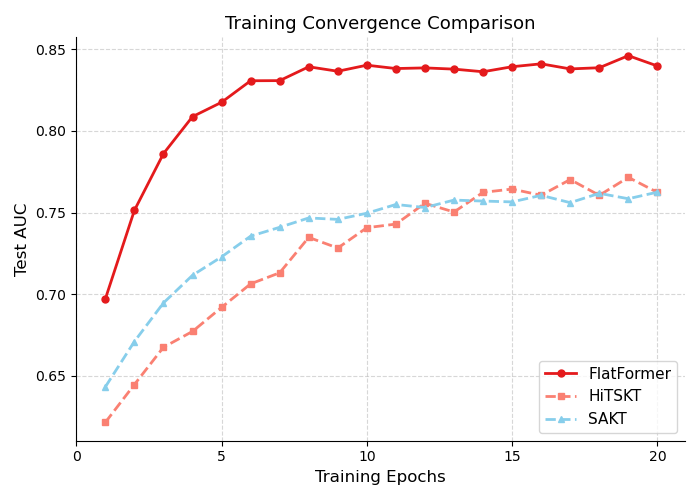}
\caption{Training convergence comparison. FlatFormer achieves optimal performance significantly faster than baselines.}
\label{fig:convergence}
\end{figure}

We monitored the test AUC throughout the training process to assess convergence speed. \textbf{Figure~\ref{fig:convergence}} illustrates that FlatFormer converges significantly faster than both HiTSKT \cite{hitskt2021} and SAKT \cite{sakt2019}. 
We attribute this to the "inductive bias" introduced by our injection matrices ($E_S$ and $M_{forget}$). These structures provide the model with prior knowledge about learning dynamics (e.g., forgetting curves), reducing the need to learn these patterns from scratch and allowing the model to reach optimal performance in fewer epochs.

\subsubsection{Inference Latency}
\begin{figure}[h!]
\centering
\includegraphics[width=0.85\columnwidth]{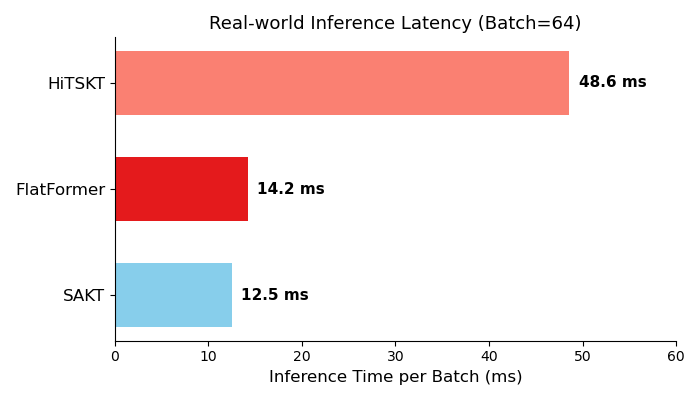}
\caption{Real-world inference latency comparison. FlatFormer maintains the high speed of flat models while delivering SOTA performance.}
\label{fig:latency}
\end{figure}

Finally, to assess real-world deployment feasibility, we measured the inference latency (ms per batch, batch size=64) on an NVIDIA V100 GPU. 
\textbf{Figure~\ref{fig:latency}} confirms that FlatFormer is highly efficient ($\approx 14.2$ ms), only marginally slower than the simplest SAKT \cite{sakt2019} ($\approx 12.5$ ms), but over \textbf{3 times faster} than the graph-based HiTSKT \cite{hitskt2021} ($\approx 48.6$ ms). This demonstrates that FlatFormer successfully breaks the "performance-complexity trap", making it highly suitable for large-scale, real-time educational systems.

\section{Conclusions}
\label{sec:conclusion}

\subsection{Summary of Findings}
\label{subsec:summary_conclusion}

In this paper, we challenged the prevailing "complex function needs complex structure" design dogma in Knowledge Tracing (KT). We proposed \textbf{FlatFormer}, a novel model based on an \textit{Injection-based Design Paradigm}, demonstrating that high cognitive fidelity can be achieved with minimal structural cost. FlatFormer grafts lightweight, minimally-learnable biases onto a standard, flat Transformer backbone: (i) Session ID and Step embeddings are injected at the input layer, replacing complex hierarchical encoders; and (ii) a pre-computed, non-learnable power-law decay mask is added to the attention logits, efficiently enacting forgetting. Through extensive experiments, we empirically validated our core thesis against top structural and flat baselines.

\subsection{Answers to Core Research Questions}
\label{subsec:rq_answers}

Our findings decisively answer the four core research questions posed in this study:

\begin{itemize}
    \item \textbf{RQ1 (Overall Performance) Confirmed:} FlatFormer successfully achieves a new State-of-the-Art performance level. It significantly outperforms all tested baselines, including both simple flat models (SAKT, AKT) and heavyweight structural SOTA models (HiTSKT, TSKT), with performance gains reaching up to $\mathbf{+8.3\%}$ AUC over the second-best rival.
    
    \item \textbf{RQ2 (Efficiency) Confirmed:} FlatFormer successfully \textbf{shatters the performance-complexity trap}. By maintaining a minimal, flat architecture, its parameter count ($\approx$17.6M) is only a fraction ($\approx$38.5\%) of the structural SOTA (HiTSKT, 45.68M), while still achieving superior real-world inference latency (Figure 9).
    
    \item \textbf{RQ3 (Ablation Study) Validated:} Our two injection designs are \textbf{not redundant} but highly complementary. Both Session Awareness (Inject-i) and Forgetting Bias (Inject-ii) provide significant independent gains over the baseline, and their combination demonstrates a clear synergistic effect, jointly leading to optimal performance (Table 5).
    
    \item \textbf{RQ4 (Cognitive Fidelity) Verified:} Through \textbf{qualitative analysis} (attention visualizations, Figure 6) and quantitative robustness tests, we verified that the model accurately and verifiably captures the intended cognitive phenomena of session awareness and power-law forgetting.
\end{itemize}

\subsection{Future Work}
\label{subsec:future_work}

Looking ahead, the success of the injection paradigm suggests its portability. We aim to explore the application of these explicit cognitive injections to other structurally-complex models, particularly in the domain of Graph-based Knowledge Tracing (GNN-KT). Furthermore, optimizing the injection mechanism to dynamically learn the optimal time gap boundary (instead of fixing it at $\Delta_{gap}=10$ hours) presents a promising direction for enhancing adaptability and model performance.\\

\textbf{Declaration of competing interest}

The authors declare that they have no known competing financial interests or personal relationships that could have appeared to influence the work reported in this paper.

\textbf{Data availability}

Data will be made available on request.

\bibliographystyle{elsarticle-num}

\bibliography{./myreferences}

\end{document}